\documentclass[10pt]{article}

\usepackage[utf8]{inputenc} 
\usepackage[T1]{fontenc}    
\usepackage{hyperref}       
\usepackage{url}            
\usepackage{booktabs}       
\usepackage{amsfonts}       
\usepackage{nicefrac}       
\usepackage{microtype}      
\usepackage{color}
\usepackage{graphicx}
\usepackage{caption}
\usepackage{subcaption}
\usepackage{float}
\usepackage{amsmath}
\usepackage{fullpage}
\usepackage[numbers,sort,compress]{natbib}

\graphicspath{{images/}}
\title{A Study of Human Gaze Behavior During Visual Crowd Counting} 

\author{%
Raji Annadi$^{1}$, Yupei Chen$^{2}$, Viresh Ranjan$^{1}$, \\ Dimitris Samaras$^{1}$, Gregory Zelinsky$^{2}$,  Minh Hoai$^{1}$ \\
$^{1}$Department of Computer Science, 
$^{2}$Department of Psychology \\
Stony Brook University, Stony Brook, NY 11790
}
\date{}

\begin{document}
\def\mA{\mathcal{A}}
\def\mB{\mathcal{B}}
\def\mC{\mathcal{C}}
\def\mD{\mathcal{D}}
\def\mE{\mathcal{E}}
\def\mF{\mathcal{F}}
\def\mG{\mathcal{G}}
\def\mH{\mathcal{H}}
\def\mI{\mathcal{I}}
\def\mJ{\mathcal{J}}
\def\mK{\mathcal{K}}
\def\mL{\mathcal{L}}
\def\mM{\mathcal{M}}
\def\mN{\mathcal{N}}
\def\mO{\mathcal{O}}
\def\mP{\mathcal{P}}
\def\mQ{\mathcal{Q}}
\def\mR{\mathcal{R}}
\def\mS{\mathcal{S}}
\def\mT{\mathcal{T}}
\def\mU{\mathcal{U}}
\def\mV{\mathcal{V}}
\def\mW{\mathcal{W}}
\def\mX{\mathcal{X}}
\def\mY{\mathcal{Y}}
\def\mZ{\mathcal{Z}}

\def\1n{\mathbf{1}_n}
\def\0{\mathbf{0}}
\def\1{\mathbf{1}}

\def\A{{\bf A}}
\def\B{{\bf B}}
\def\C{{\bf C}}
\def\D{{\bf D}}
\def\E{{\bf E}}
\def\F{{\bf F}}
\def\G{{\bf G}}
\def\H{{\bf H}}
\def\I{{\bf I}}
\def\J{{\bf J}}
\def\K{{\bf K}}
\def\L{{\bf L}}
\def\M{{\bf M}}
\def\N{{\bf N}}
\def\O{{\bf O}}
\def\P{{\bf P}}
\def\Q{{\bf Q}}
\def\R{{\bf R}}
\def\S{{\bf S}}
\def\T{{\bf T}}
\def\U{{\bf U}}
\def\V{{\bf V}}
\def\W{{\bf W}}
\def\X{{\bf X}}
\def\Y{{\bf Y}}
\def\Z{{\bf Z}}

\def\a{{\bf a}}
\def\b{{\bf b}}
\def\c{{\bf c}}
\def\d{{\bf d}}
\def\e{{\bf e}}
\def\f{{\bf f}}
\def\g{{\bf g}}
\def\h{{\bf h}}
\def\i{{\bf i}}
\def\j{{\bf j}}
\def\k{{\bf k}}
\def\l{{\bf l}}
\def\m{{\bf m}}
\def\n{{\bf n}}
\def\o{{\bf o}}
\def\p{{\bf p}}
\def\q{{\bf q}}
\def\r{{\bf r}}
\def\s{{\bf s}}
\def\t{{\bf t}}
\def\u{{\bf u}}
\def\v{{\bf v}}
\def\w{{\bf w}}
\def\x{{\bf x}}
\def\y{{\bf y}}
\def\z{{\bf z}}

\def\balpha{\mbox{\boldmath{$\alpha$}}}
\def\bbeta{\mbox{\boldmath{$\beta$}}}
\def\bdelta{\mbox{\boldmath{$\delta$}}}
\def\bgamma{\mbox{\boldmath{$\gamma$}}}
\def\blambda{\mbox{\boldmath{$\lambda$}}}
\def\bsigma{\mbox{\boldmath{$\sigma$}}}
\def\btheta{\mbox{\boldmath{$\theta$}}}
\def\bomega{\mbox{\boldmath{$\omega$}}}
\def\bxi{\mbox{\boldmath{$\xi$}}}
\def\bnu{\mbox{\boldmath{$\nu$}}}                                  
\def\bphi{\mbox{\boldmath{$\phi$}}}
\def\bmu{\mbox{\boldmath{$\mu$}}}

\def\bDelta{\mbox{\boldmath{$\Delta$}}}
\def\bOmega{\mbox{\boldmath{$\Omega$}}}
\def\bPhi{\mbox{\boldmath{$\Phi$}}}
\def\bLambda{\mbox{\boldmath{$\Lambda$}}}
\def\bSigma{\mbox{\boldmath{$\Sigma$}}}
\def\bGamma{\mbox{\boldmath{$\Gamma$}}}

\newcommand{\myminimum}[1]{\mathop{\textrm{minimum}}_{#1}}
\newcommand{\mymaximum}[1]{\mathop{\textrm{maximum}}_{#1}}    
\newcommand{\mymin}[1]{\mathop{\textrm{minimize}}_{#1}}
\newcommand{\mymax}[1]{\mathop{\textrm{maximize}}_{#1}}
\newcommand{\mymins}[1]{\mathop{\textrm{min.}}_{#1}}
\newcommand{\mymaxs}[1]{\mathop{\textrm{max.}}_{#1}}  
\newcommand{\myargmin}[1]{\mathop{\textrm{argmin}}_{#1}} 
\newcommand{\myargmax}[1]{\mathop{\textrm{argmax}}_{#1}} 
\newcommand{\myst}{\textrm{s.t. }}

\newcommand{\denselist}{\itemsep -1pt}
\newcommand{\sparselist}{\itemsep 1pt}

\definecolor{pink}{rgb}{0.9,0.5,0.5}
\definecolor{purple}{rgb}{0.5, 0.4, 0.8}   
\definecolor{gray}{rgb}{0.3, 0.3, 0.3}
\definecolor{mygreen}{rgb}{0.2, 0.6, 0.2}

\newcommand{\cyan}[1]{\textcolor{cyan}{#1}}
\newcommand{\red}[1]{\textcolor{red}{#1}}  
\newcommand{\blue}[1]{\textcolor{blue}{#1}}
\newcommand{\magenta}[1]{\textcolor{magenta}{#1}}
\newcommand{\pink}[1]{\textcolor{pink}{#1}}
\newcommand{\green}[1]{\textcolor{green}{#1}} 
\newcommand{\gray}[1]{\textcolor{gray}{#1}}    
\newcommand{\mygreen}[1]{\textcolor{mygreen}{#1}}    
\newcommand{\purple}[1]{\textcolor{purple}{#1}}       

\definecolor{greena}{rgb}{0.4, 0.5, 0.1}
\newcommand{\greena}[1]{\textcolor{greena}{#1}}

\definecolor{bluea}{rgb}{0, 0.4, 0.6}
\newcommand{\bluea}[1]{\textcolor{bluea}{#1}}
\definecolor{reda}{rgb}{0.6, 0.2, 0.1}
\newcommand{\reda}[1]{\textcolor{reda}{#1}}

\def\changemargin#1#2{\list{}{\rightmargin#2\leftmargin#1}\item[]}
\let\endchangemargin=\endlist
                                               
\newcommand{\cm}[1]{}

\newcommand{\mtodo}[1]{{\color{red}$\blacksquare$\textbf{[TODO: #1]}}}
\newcommand{\myheading}[1]{\vspace{1ex}\noindent \textbf{#1}}
\newcommand{\htimesw}[2]{\mbox{$#1$$\times$$#2$}}
\newcommand{\mh}[1]{\textcolor{blue}{[Minh: {#1}]}}
\newcommand{\ms}[1]{\textcolor{red}{[MS: {#1}]}}
\newcommand{\yc}[1]{\textcolor{purple}{[Yupei: {#1}]}}
\newcommand{\vr}[1]{\textcolor{green}{[Viresh: {#1}]}}

\newif\ifshowsolution
\showsolutiontrue

\ifshowsolution  
\newcommand{\Comment}[1]{\paragraph{\bf $\bigstar $ COMMENT:} {\sf #1} \bigskip}
\newcommand{\Solution}[2]{\paragraph{\bf $\bigstar $ SOLUTION:} {\sf #2} }
\newcommand{\Mistake}[2]{\paragraph{\bf $\blacksquare$ COMMON MISTAKE #1:} {\sf #2} \bigskip}
\else
\newcommand{\Solution}[2]{\vspace{#1}}
\fi

\newcommand{\truefalse}{
\begin{enumerate}
	\item True
	\item False
\end{enumerate}
}

\newcommand{\yesno}{
\begin{enumerate}
	\item Yes
	\item No
\end{enumerate}
}
\newcommand{\Sref}[1]{Sec.~\ref{#1}}
\newcommand{\Eref}[1]{Eq.~(\ref{#1})}
\newcommand{\Fref}[1]{Fig.~\ref{#1}}
\newcommand{\Tref}[1]{Tab.~\ref{#1}}

 
\maketitle

\begin{abstract}
In this paper, we describe our study on how humans allocate their attention during visual crowd counting. Using an eye tracker, we collect gaze behavior of human participants who are tasked with counting the number of people in crowd images. Analyzing the collected gaze behavior of ten human participants on thirty crowd images, we observe some common approaches for visual counting. For an image of a small crowd, the  approach is to enumerate over all people or groups of people in the crowd, and this explains the high level of similarity between the fixation density maps of different human participants. For an image of a large crowd, our participants tend to focus on one section of the image, count the number of people in that section, and then extrapolate to the other sections. In terms of count accuracy, our human participants are  not as good at the counting task, compared to the performance of the current state-of-the-art computer algorithms. Interestingly, there is a tendency to under count the number of people in all crowd images. Gaze behavior data and images can be downloaded from 
\url{https://www3.cs.stonybrook.edu/~minhhoai/projects/crowd_counting_gaze/}.

\end{abstract}

\section{Introduction}


 Large crowd gathering is commonplace, and estimating the size of a crowd is an important problem for different purposes ranging from journalism to public safety. It is known that humans are good at subitizing~\cite{Kaufman-etal-AJP49}, i.e., predicting fast and accurate counts for small number of items. For an image of a dense crowd, humans can also provide a very accurate count given enough time. In fact, one common approach to produce `gold standard' annotations \cite{idrees2018composition, wang2020nwpu, zhang2016single, idrees2013multi}  for training computer algorithms ~\cite{ li2018csrnet,ranjan2018iterative,idrees2018composition,cao2018scale,bayesianCounting,wang2019learning,liu2019context,shi2019revisiting,liu2019point,liu2019adcrowdnet,Xu_2019_ICCV,Cheng_2019_ICCV,Liu_2019_ICCV,Yan_2019_ICCV,Zhao_2019_CVPR,Zhang_2019_CVPR,Jiang_2019_CVPR,Wan_2019_CVPR,Lian_2019_CVPR,Liu_2019_CVPR, lu2018class, wan2019adaptive, m_Ranjan-etal-ACCV20, m_Wang-etal-NIPS20}  for crowd counting is to ask human annotators to enumerate and place a dot on each person in the crowd images. The things that remain unclear are: 1) how well can humans count a large number of objects in dense images under limited time, and 2) what are the approaches that humans typically use for counting when there is not enough time to attend to and enumerate over all objects. In this paper, we seek the answers to these two questions by collecting and analyzing the gaze behavior of human subjects that were asked to count under a time constraint.






\section{Data Collection}

 The gaze-behavior dataset we collected consists of 23,036 fixations from 10 participants with the task of counting the number of people in 30 images for 30 seconds per image. The 10 participants were Stony Brook University undergraduate and graduate students (6 male, 4 female, age range 18--28, normal or corrected-to-normal vision). A total of 30 diverse images were selected from the UCF-QNRF dataset~\cite{idrees2018composition} and used in this data collection. These images contained as few as 65 people and as many as 12672 people  and included both images taken in daylight and at night.

Each participant was instructed to freely view each image and count ``how many people were there in the image?'' Each trial started with a central fixation dot and the participant had to fixate at the dot and press a button to start the trial. When the image appeared, they had 30s to freely inspect the image and count the number of people in that image. The image disappeared after 30 seconds and they were asked to type in a number that indicated the number of people in the image. The 30 images were randomly interleaved and it took $\sim$20 minutes to finish the experiment. Eye movements were recorded using an EyeLink 1000 eye-tracker in desktop configuration (SR research) with a sample rate at 1000Hz. Eye-tracker calibration was conducted at the beginning of the experiment. All participants completed the experiment in the same controlled laboratory condition.




\section{Gaze Behavior Analysis}
\label{headings}

This section describes several notable patterns in the gaze behaviors of our ten participants, obtained by visualizing their fixation density maps. All fixation density maps of the 10 participants on 30 crowd images are shown in the last 30 figures of this paper.

\begin{figure}[h]
\centering 
\begin{subfigure}{0.47\textwidth}
\includegraphics[width=1\linewidth, height=0.7\linewidth]{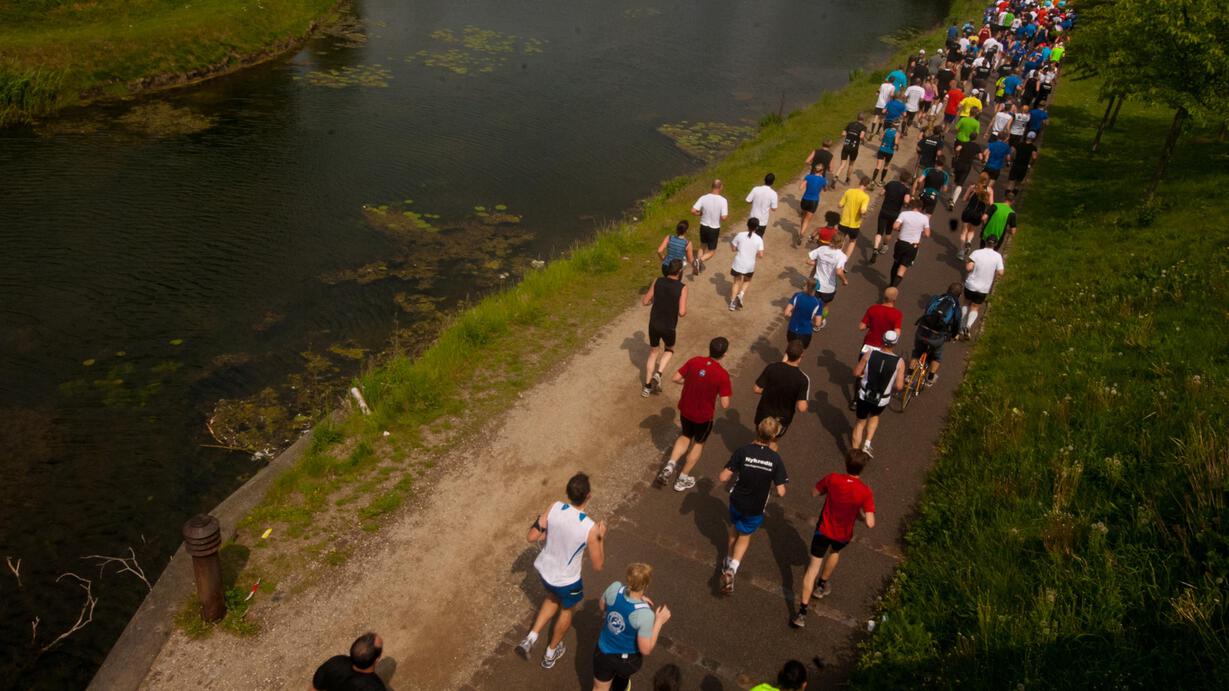} 
\caption{A sparse crowd image}
\label{fig:fig1a}
\end{subfigure}
\begin{subfigure}{0.47\textwidth}
\includegraphics[width=1\linewidth, height=0.7\linewidth]{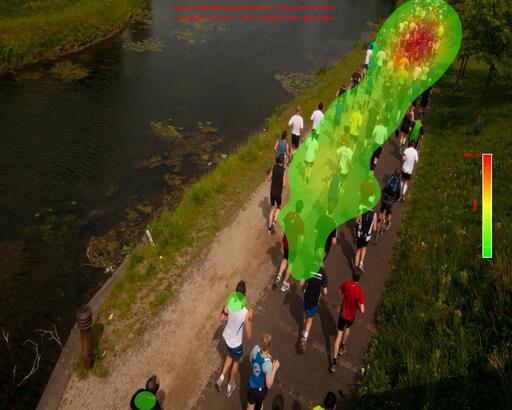}
\caption{Fixation density map}
\label{fig:fig1b}
\end{subfigure}
\caption{A sparse crowd image and the corresponding fixation density map}
\label{fig:fig1}
\end{figure}

\begin{figure}[h]
\centering 
\begin{subfigure}{0.47\textwidth}
\includegraphics[width=1\linewidth, height=0.7\linewidth]{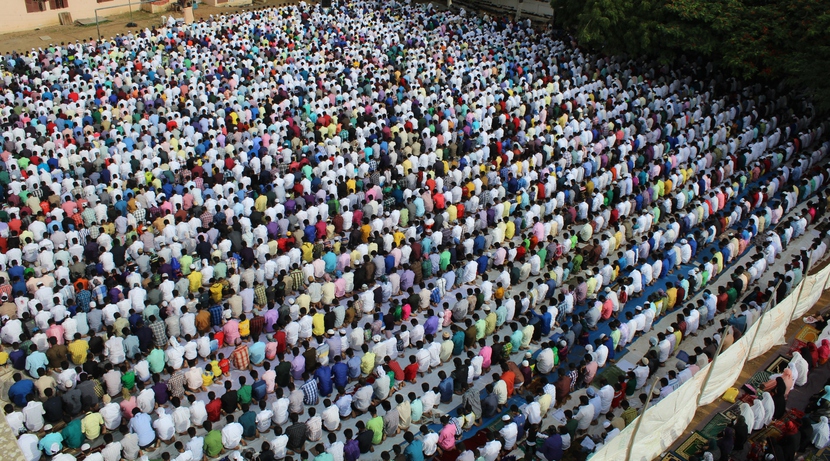} 
\caption{A dense crowd image}
\label{fig:fig2a}
\end{subfigure}
\begin{subfigure}{0.47\textwidth}
\includegraphics[width=1\linewidth, height=0.7\linewidth]{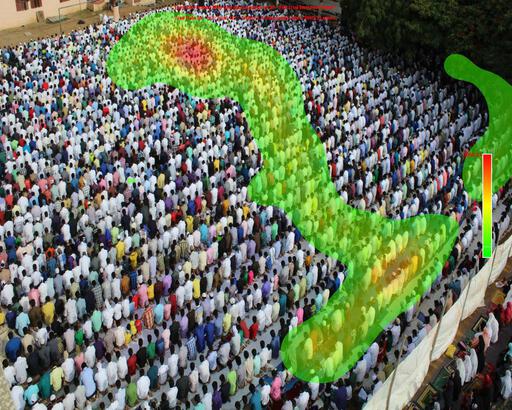}
\caption{Fixation density map}
\label{fig:fig2b}
\end{subfigure}

\caption{A dense crowd image and the corresponding fixation density map}
\label{fig:image2}
\end{figure}

Our first observation is that our participants use different approaches for images of sparse and dense crowds. For an image with a sparse crowd where the total number of people is small (less than approximately $100$ people) such as the one in Figure~\ref{fig:fig1a}, the fixation density map covers the entire crowd area as can be seen in Figure~\ref{fig:fig1b}. This suggests that the participants pay attention to every area of the crowd, and they probably aim to account for every single person or group of people in the image. For an image with a denser crowd such as the one in Figure~\ref{fig:fig2a}, the fixation density map covers only a small part of the crowd. This is understandable given the size of the crowd and the impossibility of enumerating over all people in the crowd. For an image of a dense crowd, the fixation density maps reveal that our participants tend to focus on one section of the crowd, count or estimate the number of people in that section, and  extrapolate to the other sections. 

\begin{figure}
    \centering
    \includegraphics[width=0.8\linewidth]{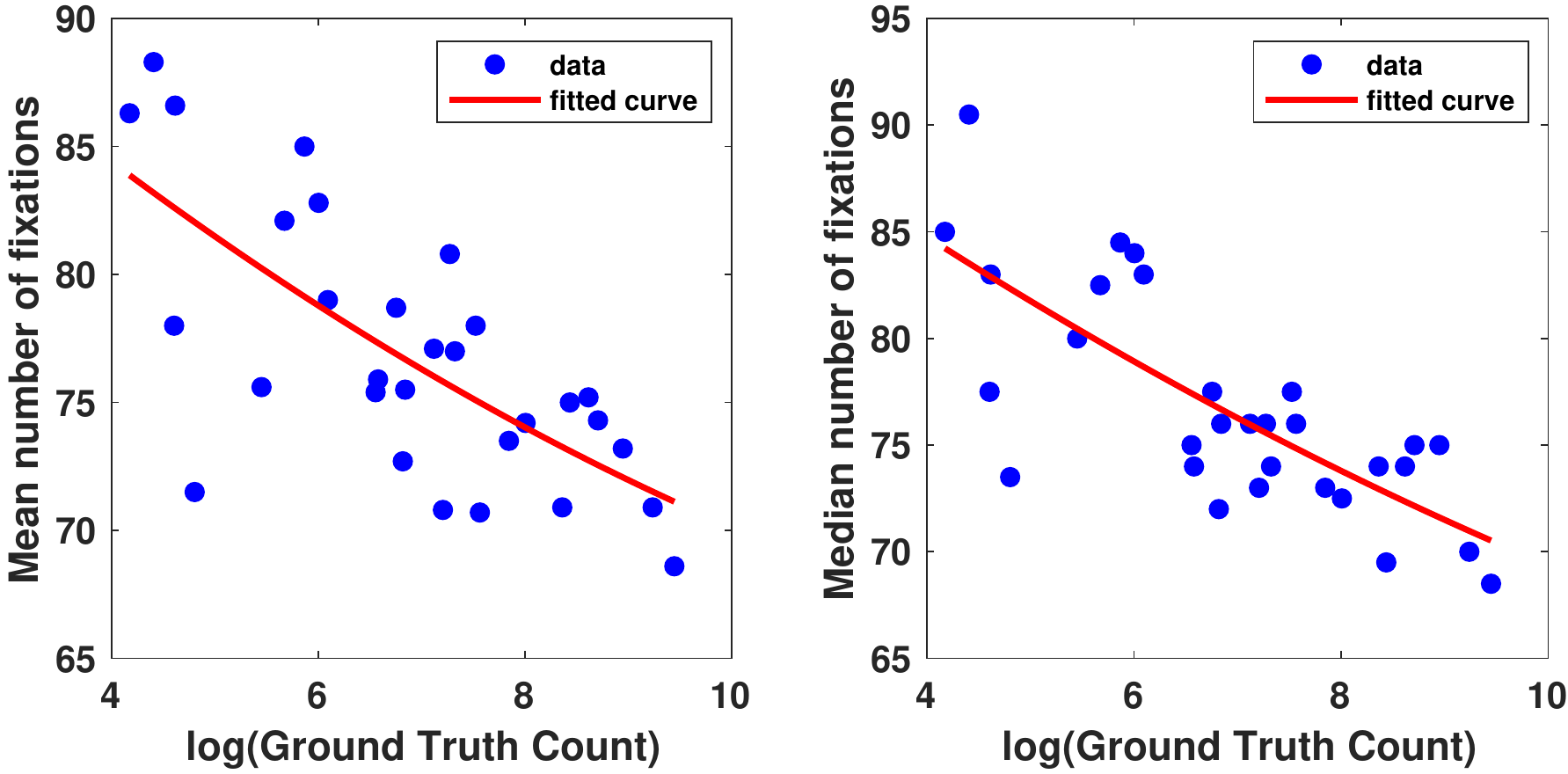}
    \caption{The mean and median number of fixations as a function of ground truth count. The number of fixations tends to decrease as the size of the crowd increases. }
    \label{fig:fig3}
\end{figure}

Our second observation is that the number of fixations tends to decrease as the size of the crowd increases, as can be seen in Figure~\ref{fig:fig3}. For a sparse crowd, our participants were fast to change their eye fixations, and this indicates an enumeration behavior. Meanwhile for a dense crowd, there is some region of the image where our participants attended to for a longer period of time to estimate the total count in the region. 


\begin{figure}
     \centering
         \includegraphics[width=0.47\textwidth]{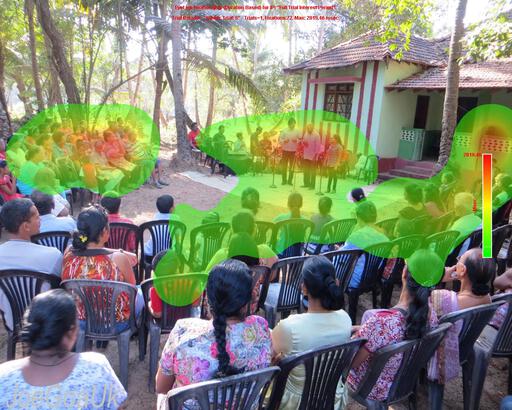}
         \includegraphics[width=0.47\textwidth]{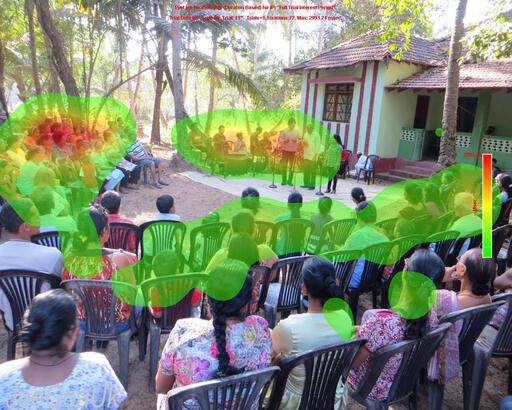}
     
        \caption{Fixation density maps of two different subjects on the same image with a sparse crowd. These density maps are highly similar to each other.}
        \label{fig:similar_fdm}
\end{figure}

Our third observation is that there is high degree of similarity between the fixation density maps of different subjects for an input image of a sparse crowd, as shown in Figure~\ref{fig:similar_fdm}. This further indicates the behavior to account for all people or group of people in a sparse crowd. 


\begin{figure}
     \centering
        \includegraphics[width=0.32\textwidth]{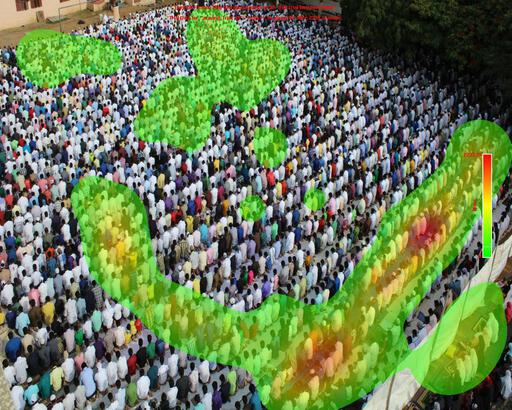}
         \includegraphics[width=0.32\textwidth]{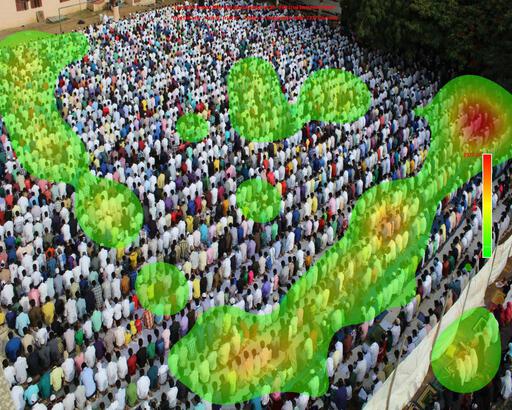}
         \includegraphics[width=0.32\textwidth]{fix_map_dense}
        \caption{Fixation density maps from three different subjects on a crowd with a uniform  density. The common behavior is to count the number of people in one row and count the number of rows. }
        \label{fig:fig4}
\end{figure}

We also find that our participants are more successful in estimating the size of a crowd with a uniform spread. Their common approach is to count the number of people in a selected row and selected column and multiple the resulting numbers to obtain the final count.  This behavior can be observed in Figure~\ref{fig:fig4}, where each fixation density map spreads over one row and one column.


\section{Count Accuracy}

\setlength{\tabcolsep}{2pt}
\begin{table}[h!]
    \centering
    \begin{tabular}{l|c|rrrrrrrrrr|c|c}
    \toprule
    & GT& \multicolumn{10}{c|}{Predicted counts of subjects} & Mean & Median\\
    & Count & Subj1 & Subj2 & Subj3 & Subj4 & Subj5 & Subj6 &  Subj7 & Subj8 & Subj9 & Subj10 & NAE-I & NAE-I\\ 
    \midrule
Image 07 & 65 & 59 & 80 & 55 & 60 & 65 & 65 & 76 & 60 & 62 & 60 & 0.09 & 0.08\\  
Image 08 & 82 & 90 & 70 & 90 & 45 & 55 & 40 & 65 & 50 & 62 & 90 & 0.26 & 0.23\\  
Image 03 & 100 & 130 & 60 & 90 & 50 & 95 & 50 & 800 & 35 & 100 & 200 & 1.05 & 0.45\\  
Image 06 & 101 & 68 & 80 & 110 & 60 & 160 & 50 & 150 & 60 & 65 & 170 & 0.40 & 0.41\\  
Image 17 & 122 & 75 & 65 & 70 & 60 & 50 & 70 & 135 & 50 & 75 & 140 & 0.40 & 0.43\\  
Image 09 & 233 & 230 & 100 & 110 & 180 & 300 & 150 & 320 & 120 & 120 & 220 & 0.34 & 0.36\\  
Image 15 & 291 & 220 & 300 & 120 & 100 & 70 & 120 & 350 & 70 & 160 & 350 & 0.45 & 0.52\\  
Image 10 & 353 & 350 & 250 & 175 & 200 & 350 & 180 & 350 & 150 & 200 & 500 & 0.32 & 0.42\\  
Image 12 & 405 & 550 & 250 & 200 & 200 & 350 & 200 & 370 & 60 & 120 & 500 & 0.43 & 0.44\\  
Image 22 & 443 & 250 & 350 & 200 & 400 & 300 & 400 & 500 & 200 & 150 & 500 & 0.32 & 0.27\\  
Image 27 & 703 & 430 & 250 & 150 & 300 & 350 & 200 & 350 & 150 & 160 & 300 & 0.62 & 0.61\\  
Image 14 & 720 & 350 & 600 & 250 & 400 & 5000 & 500 & 600 & 120 & 400 & 700 & 0.95 & 0.44\\  
Image 30 & 858 & 700 & 200 & 225 & 350 & 250 & 300 & 400 & 150 & 130 & 600 & 0.61 & 0.68\\  
Image 11 & 915 & 750 & 500 & 375 & 150 & 1000 & 400 & 10000 & 150 & 200 & 500 & 1.47 & 0.58\\  
Image 02 & 936 & 580 & 400 & 300 & 500 & 3000 & 1500 & 5000 & 150 & 480 & 5000 & 1.49 & 0.64\\  
Image 20 & 1237 & 800 & 1000 & 275 & 700 & 600 & 1800 & 1800 & 400 & 500 & 3500 & 0.63 & 0.49\\  
Image 29 & 1349 & 2000 & 550 & 400 & 1000 & 5000 & 1200 & 1900 & 500 & 500 & 900 & 0.69 & 0.54\\  
Image 05 & 1442 & 1600 & 2000 & 375 & 800 & 7000 & 1000 & 8000 & 100 & 400 & 4000 & 1.38 & 0.73\\  
Image 01 & 1515 & 900 & 1000 & 350 & 1300 & 700 & 1000 & 800 & 500 & 900 & 1500 & 0.41 & 0.41\\  
Image 28 & 1852 & 1400 & 500 & 350 & 800 & 17000 & 600 & 7000 & 600 & 250 & 4000 & 1.67 & 0.77\\  
Image 13 & 1929 & 1200 & 600 & 500 & 600 & 14 & 1500 & 9000 & 100 & 500 & 20000 & 1.84 & 0.74\\  
Image 23 & 2555 & 1300 & 1300 & 400 & 800 & 15500 & 1500 & 18000 & 300 & 1200 & 15000 & 2.03 & 0.77\\  
Image 16 & 3002 & 2000 & 4000 & 600 & 3000 & 10000 & 1600 & 8000 & 700 & 1500 & 2000 & 0.75 & 0.48\\  
Image 26 & 4282 & 1300 & 5000 & 650 & 1500 & 12000 & 2500 & 18000 & 800 & 1200 & 130000 & 3.87 & 0.77\\  
Image 18 & 4609 & 4800 & 3000 & 700 & 2000 & 18000 & 1900 & 25000 & 700 & 3000 & 15000 & 1.32 & 0.72\\  
Image 04 & 5520 & 3500 & 4000 & 1200 & 1000 & 10000 & 3000 & 15000 & 5000 & 1300 & 25000 & 0.96 & 0.77\\  
Image 24 & 6052 & 2100 & 6000 & 700 & 5000 & 20000 & 3000 & 12400 & 1000 & 1000 & 12000 & 0.82 & 0.83\\  
Image 25 & 7695 & 11000 & 10000 & 1400 & 1000 & 15000 & 5000 & 40000 & 2500 & 5000 & 28000 & 1.16 & 0.75\\  
Image 21 & 10274 & 6500 & 6000 & 900 & 1500 & 12500 & 5000 & 80000 & 2000 & 4500 & 1000000 & 10.78 & 0.68\\  
Image 19 & 12672 & 1600 & 1500 & 1200 & 4000 & 18000 & 9000 & 100000 & 2000 & 2500 & 1600000 & 13.79 & 0.86\\  
\midrule 
MAE &  & 1149 & 1050 & 1994 & 1475 & 3670 & 1025 & 9849 & 1785 & 1519 & 93453 & & \\  
RMSE &  & 2457 & 2327 & 3443 & 2836 & 5940 & 1650 & 22173 & 3081 & 2663 & 342365 & & \\  
MeanNAE &  & 0.34 & 0.39 & 0.64 & 0.49 & 1.54 & 0.43 & 2.42 & 0.68 & 0.55 & 9.63 & & \\  
MedNAE &  & 0.36 & 0.37 & 0.74 & 0.50 & 0.65 & 0.46 & 0.79 & 0.76 & 0.55 & 0.63 & & \\
\bottomrule 
\end{tabular}
    \caption{Ground truth and estimated counts from 10 different subjects on the 30 images. The images in the table are listed based on the sizes of the crowds. The table reports the Mean Absolute Error (MAE) and Root Mean Squared Error (RMSE) of individual subjects. This table also reports the Mean and Median Normalized Absolute Error (NAE) calculated for individual subjects (i.e., aggregating over images, referred as MeanNAE and MedNAE in the table) or individual images (i.e., aggregating over subjects, referred to as MeanNAE-I and MedNAE-I in the table).  }
    \label{tab:count_est}
\end{table}

Table~\ref{tab:count_est} lists the estimated counts from the 10 subjects on all 30 images. The images in the table are listed based on the sizes of the crowds, from smallest to largest. The table also reports the counting accuracy of the human subjects under several performance metrics. Following previous works on automated crowd counting (e.g.,~\cite{ranjan2018iterative,idrees2018composition,cao2018scale,idrees2013multi,zhang2016single}), we  evaluate the accuracy of human counting based on: Mean Absolute Error (MAE), Root Mean Squared Error (RMSE), Mean Normalized Absolute Error (MeanNAE), and Median Normalized Aboslute Error (MedNAE). Let $C_i$ denote the ground truth count for Image $i$, and $\hat{C}_{i}^j$ is the estimated count from Subject $j$ for Image $i$. The MAE, RMSE, MeanNAE and MedMAE for Subject $j$ are defined as follows.
\begin{align}
&\textrm{MAE} = \frac{1}{n}\sum_{i=1}^{n} \lvert C_i - \hat{C}_i^j \rvert; \quad  
\textrm{RMSE} = \sqrt[]{\frac{1}{n}\sum_{i=1}^{n} (C_i - \hat{C}_i^j)^2}; \quad \\
&\textrm{MeanNAE} = \frac{1}{n}\sum_{i=1}^{n} \frac{\lvert C_i - \hat{C}_i^j \rvert}{C_i}; \quad \textrm{MedNAE} = \mathop{\textrm{Median}}_{i=\overline{1,n}} \left\{\frac{\lvert C_i - \hat{C}_i^j \rvert}{C_i}\right\},
\end{align}
where $n=30$ the number of images. We also compute the Mean NAE and Median NAE for each individual image by taking the mean or median over the 10 subjects. For Image $i$, the Mean NAE (MeanNAE-I) and Median NAE (MedNAE-I) are defined as follows:
\begin{align}
&\textrm{MeanNAE-I} = \frac{1}{m}\sum_{j=1}^{m} \frac{\lvert C_i - \hat{C}_i^j \rvert}{C_i}; \quad \textrm{MedNAE-I} = \mathop{\textrm{Median}}_{j=\overline{1,m}} \left\{\frac{\lvert C_i - \hat{C}_i^j \rvert}{C_i}\right\},
\end{align}
where $m = 10$ is the number of subjects.

\begin{figure}
    \centering
    \includegraphics[width=0.7\linewidth]{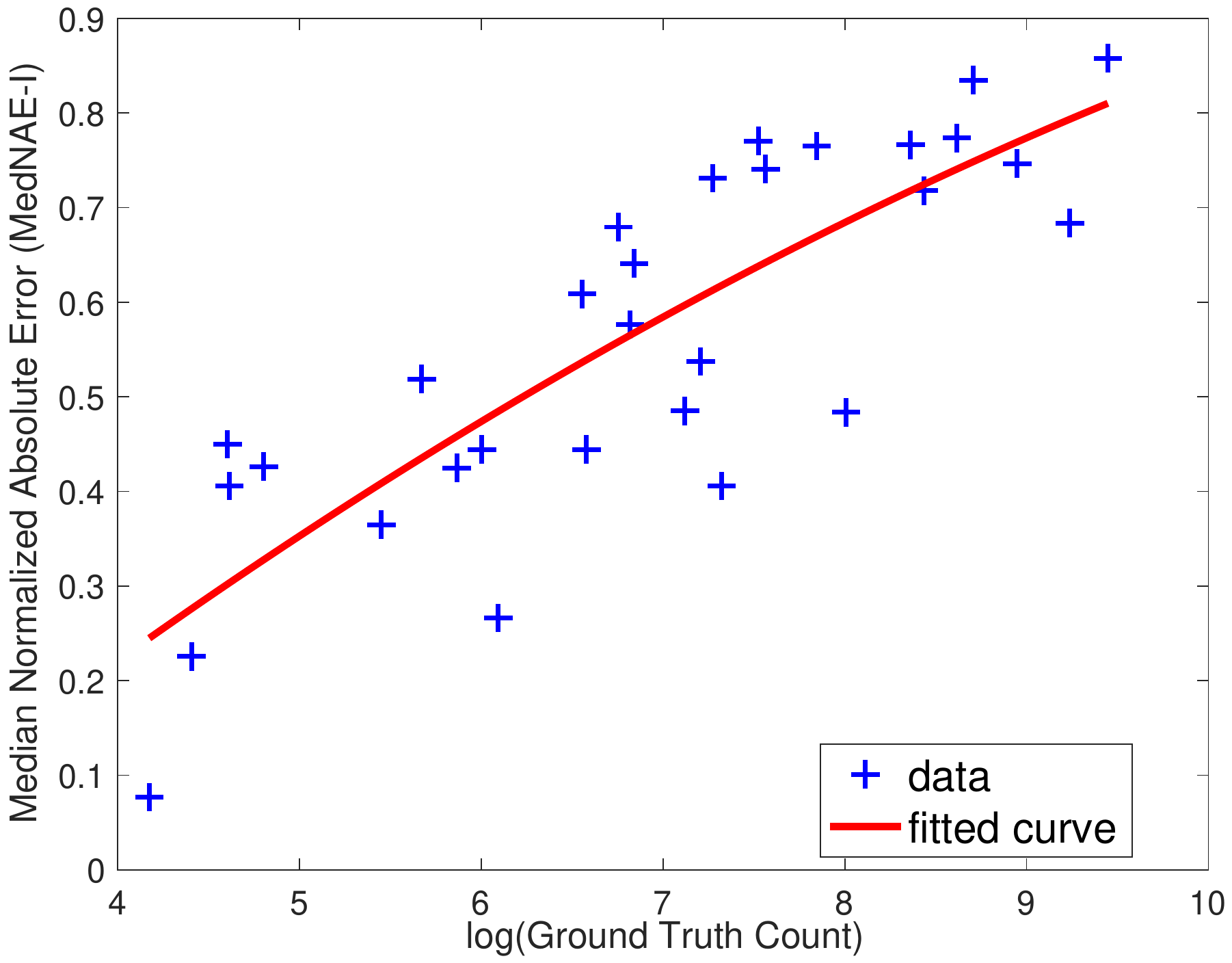}
    \caption{Median Normalized Absolute Error (MedNAE-I) versus the ground truth count in log scale. The MedNAE-I is somewhat linearly proportional to the logarithm of the crowd size. }
    \label{fig:mednae-i}
\end{figure}


\label{others}

As can be seen from Table~\ref{tab:count_est}, our human subjects are  not as good at counting under time constraint, relative to the current state-of-the-art automated crowd counting methods. The lowest MAE and RMSE values are 1025 and 1650, respectively. As a reference, the MAE and RMSE of the DMCount method~\cite{m_Wang-etal-NIPS20} on all test images of UCF-QNRF~\cite{idrees2018composition} are 85.6 and 148.3 respectively.

Comparing between images, we see that the absolute counting error is proportional to the size of crowd. However, the correlation between the absolute error and the crowd size is not linear. Considering the absolute error normalized by  crowd size, the median value (MedNAE-I) is generally higher for a larger crowd, as can be seen from the last column of Table~\ref{tab:count_est} and Figure~\ref{fig:mednae-i}.

\begin{figure}
    \centering
    \includegraphics[width=0.6\linewidth]{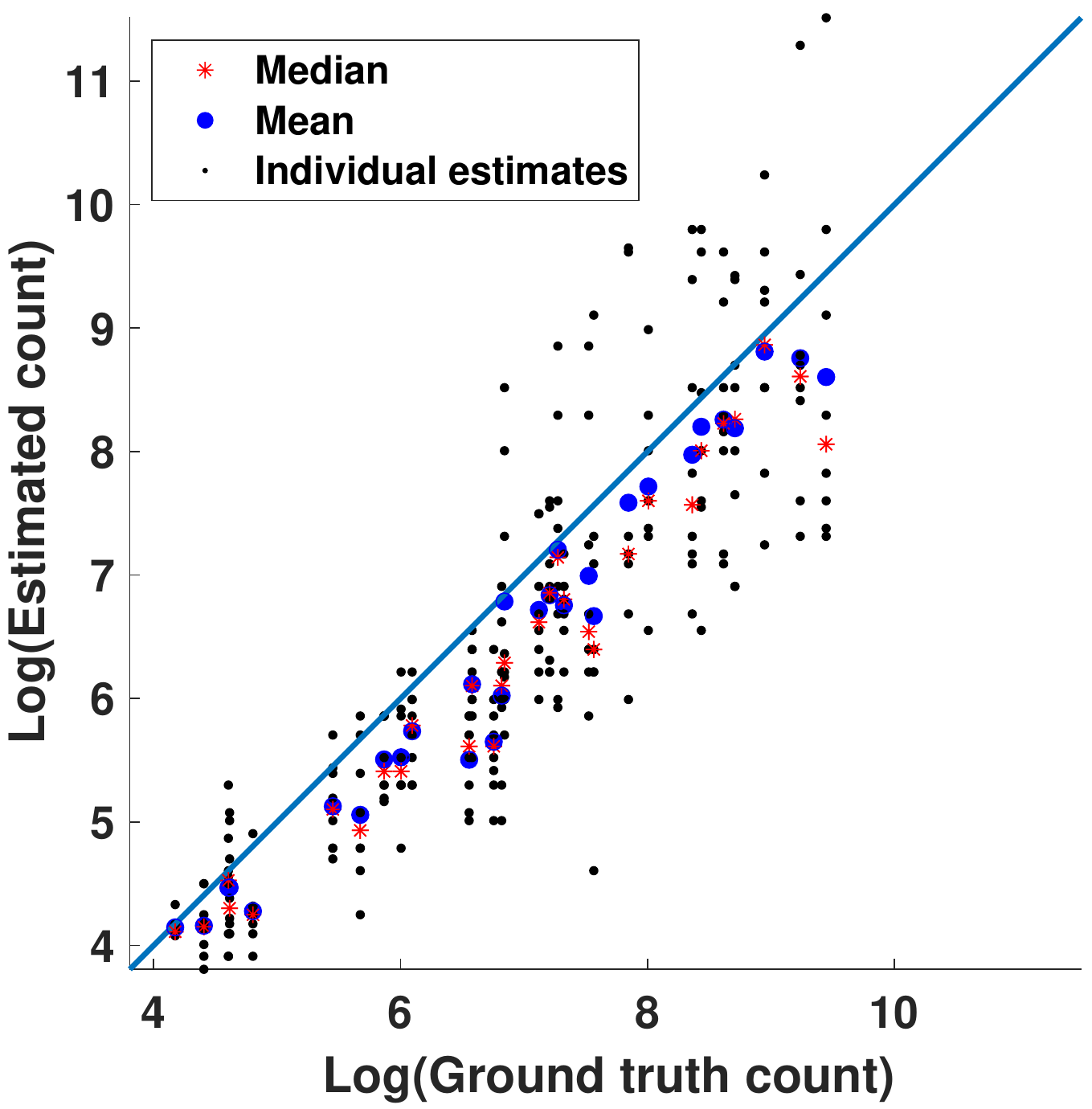}
    \caption{Estimated counts versus ground truth counts in log-scale plot. Black dots are the individual estimates from out subjects. For each image, the figure only shows eight out of ten estimates, removing the minimum and maximum estimated values; this corresponds to removing the possible outliers outside the 10 to 90 percentiles range. The blue circles and red stars are the mean and median values of the estimated values respectively. The estimates above the diagonal line correspond to over counting, while the ones below the diagonal like correspond to under counting. For all images, both mean and median values are lower the the actual counts. This indicates the under-counting tendency of our subjects. For images with a smaller number of people, most estimated counts are lower than the actual number of people in the image.  }
    \label{fig:pred_gt_plot}
\end{figure}

Figure~\ref{fig:pred_gt_plot} is the scatter plot of the estimated counts and the ground truth counts. As can be seen, our human subjects tend to under-count the number of people in each image. On all images, the median and mean estimates are smaller than the ground truth. For images with smaller crowd sizes, the majority of the estimates are lower than the actual crowd sizes. 

\section{Conclusions}
In this paper, we have presented a gaze-behavior dataset consisting of 23K fixations from 10 human participants on thirty crowd images. We have analyzed the gaze behavior of the human participants as well as the accuracy of their count estimates. We found that our human participants used different approaches for sparse and dense crowds, and the number of fixations tends to decrease as the size o the crowd increases. For a sparse crowd, the common approach of our participants was to  enumerate over all people or group of people in the crowd, leading to fixation density maps that were highly similar to one another. For a denser crowd, our participants tended to focus on counting the number of people in a small section and then extrapolate to other sections. In terms of count accuracy, our human participants performed poorly, compared to the performance of the state-of-the-art computer algorithms. We also observed a tendency to under count the number of people in each image. 







{\small
\bibliographystyle{abbrv}
\bibliography{longstrings,pubs,egbib}
}

\iftrue
\newpage 


\begin{figure}
    \centering
    \includegraphics[width=0.9\linewidth]{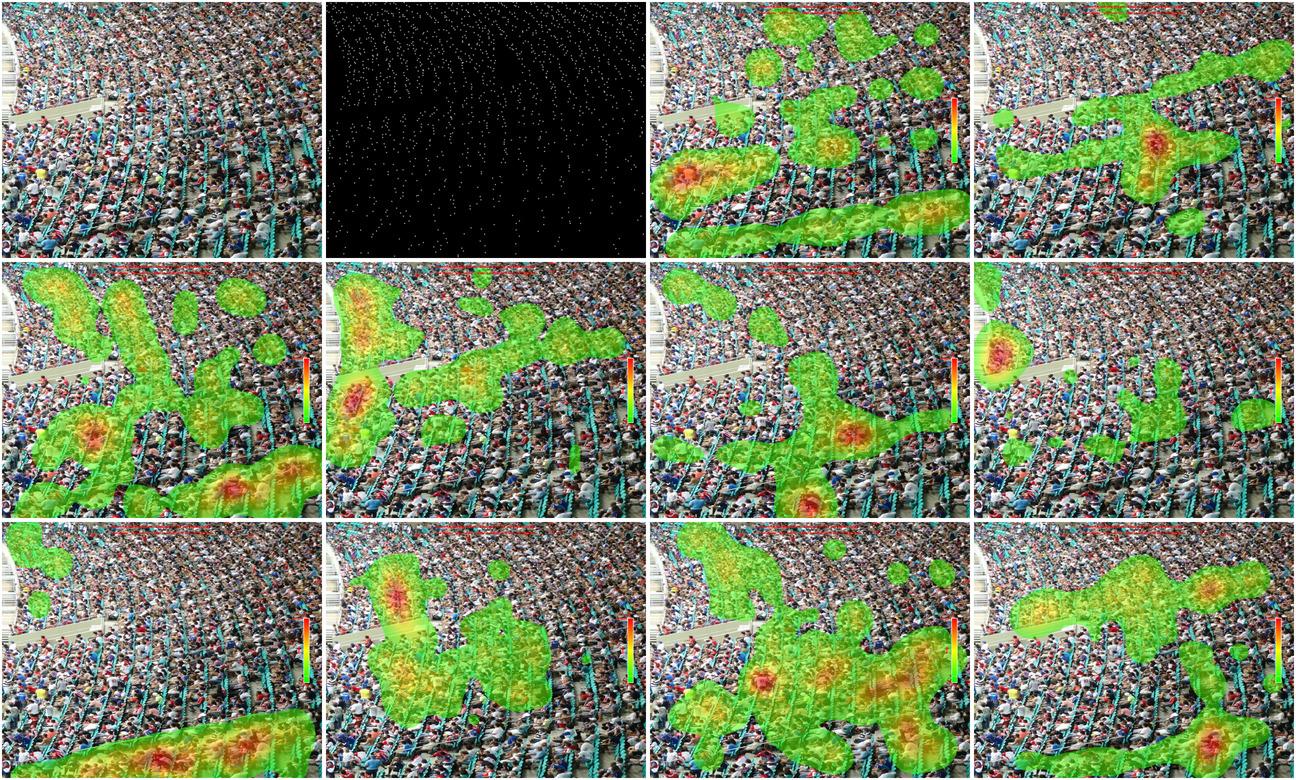}
    \caption{Image grid containing image 1, its annotation and fixation density maps  from all subjects. }
    \label{fig:image2}
\end{figure}

\begin{figure}
    \centering
    \includegraphics[width=0.9\linewidth]{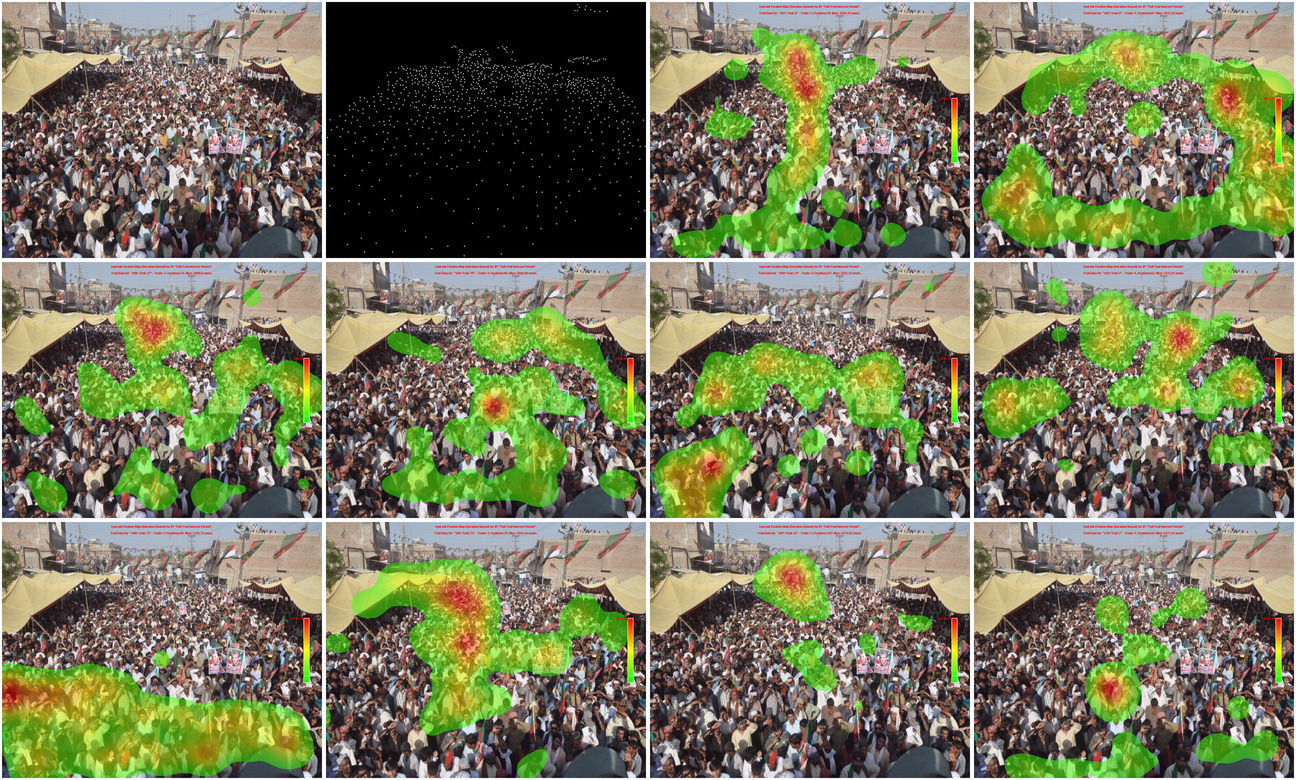}
    \caption{Image grid containing image 2, its annotation and fixation density maps  from all subjects. }
    \label{fig:image2}
\end{figure}

\begin{figure}
    \centering
    \includegraphics[width=0.9\linewidth]{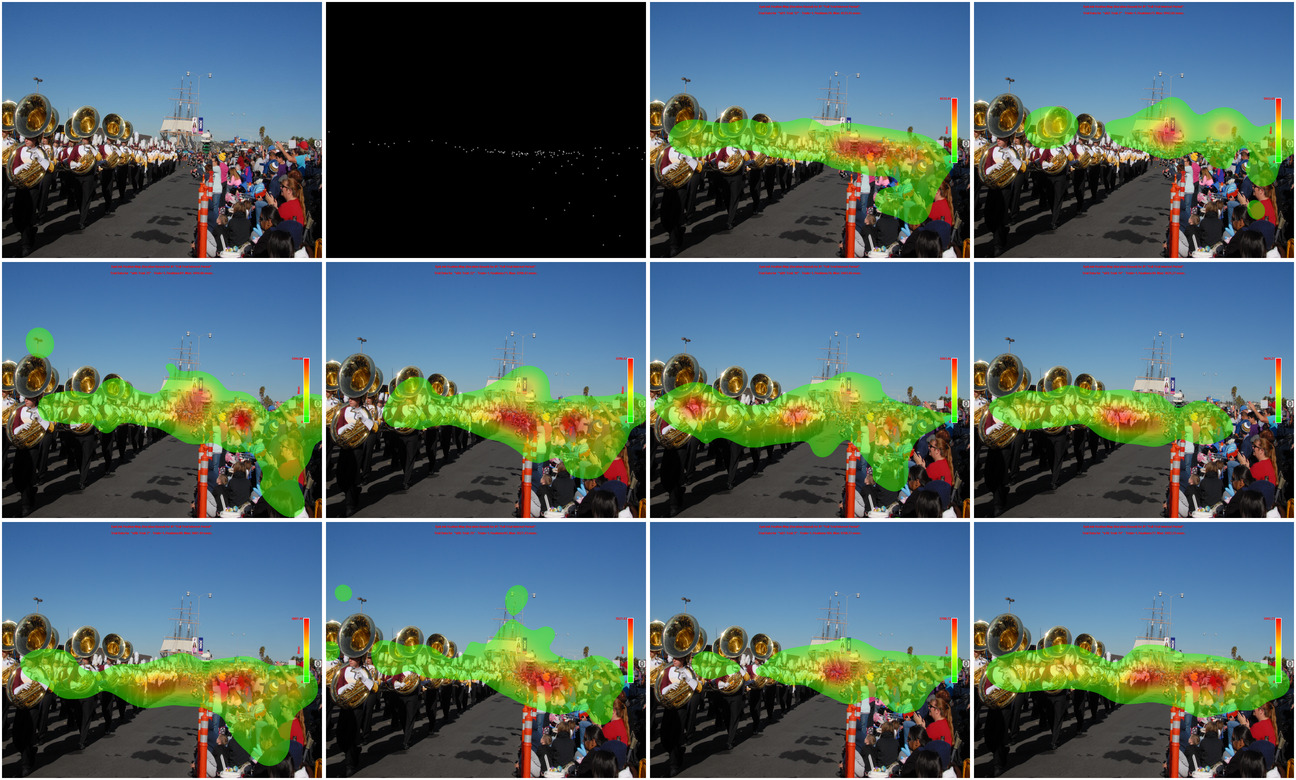}
    \caption{Image grid containing image 3, its annotation and fixation density maps  from all subjects. }
    \label{fig:image2}
\end{figure}

\begin{figure}
    \centering
    \includegraphics[width=0.9\linewidth]{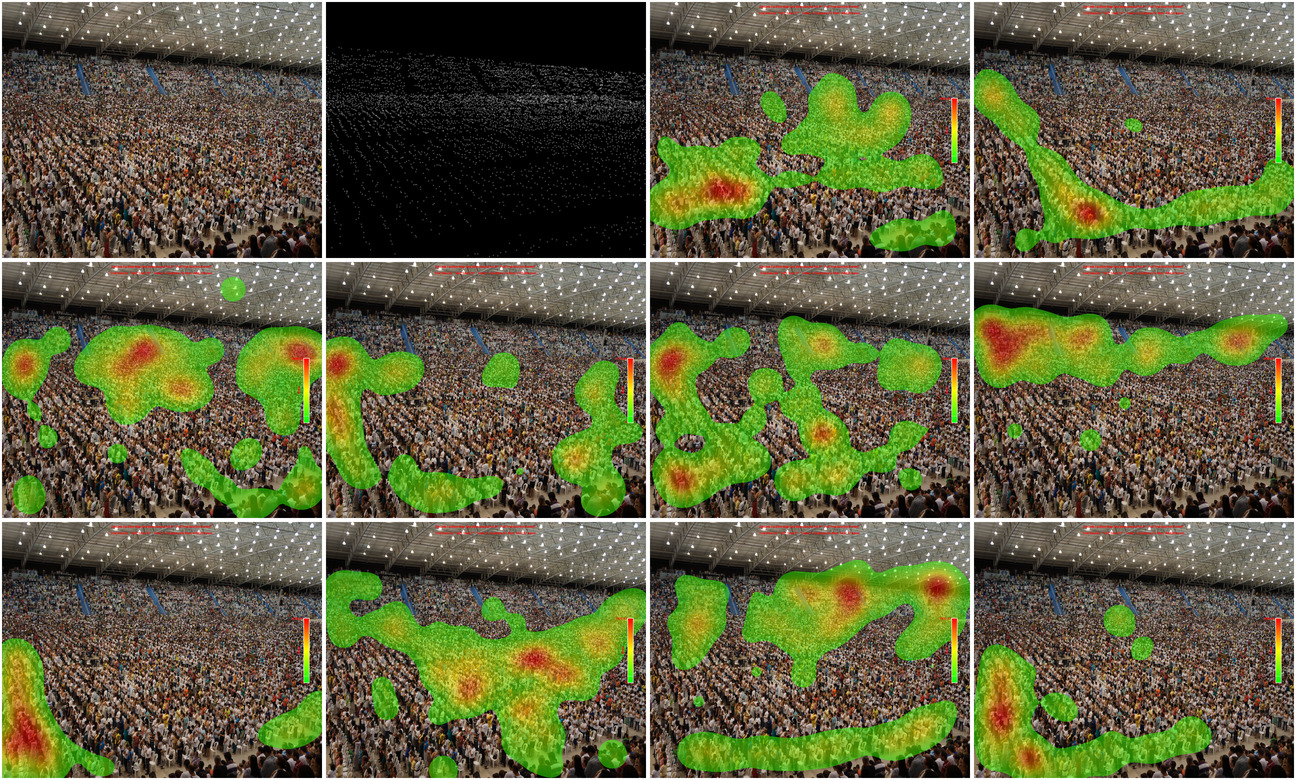}
    \caption{Image grid containing image 4, its annotation and fixation density maps  from all subjects. }
    \label{fig:image2}
\end{figure}

\begin{figure}
    \centering
    \includegraphics[width=0.9\linewidth]{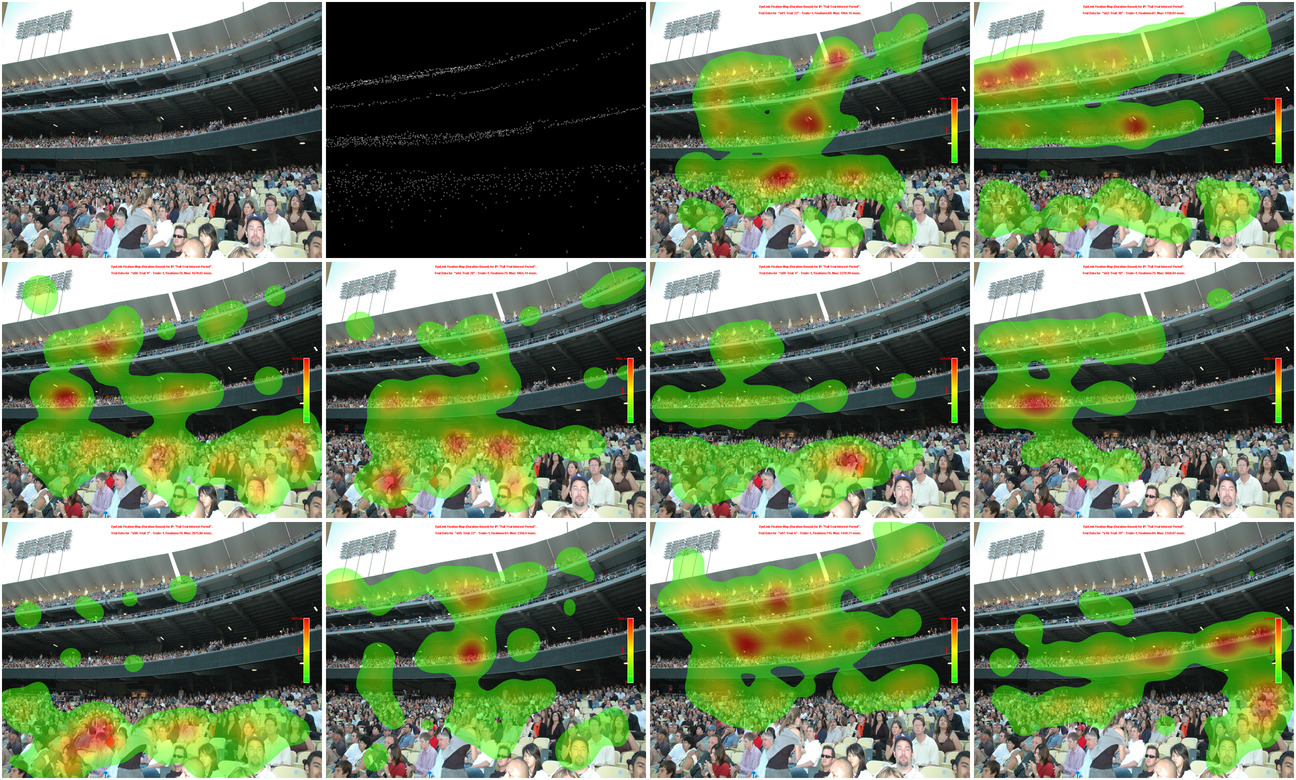}
    \caption{Image grid containing image 5, its annotation and fixation density maps  from all subjects. }
    \label{fig:image2}
\end{figure}

\begin{figure}
    \centering
    \includegraphics[width=0.9\linewidth]{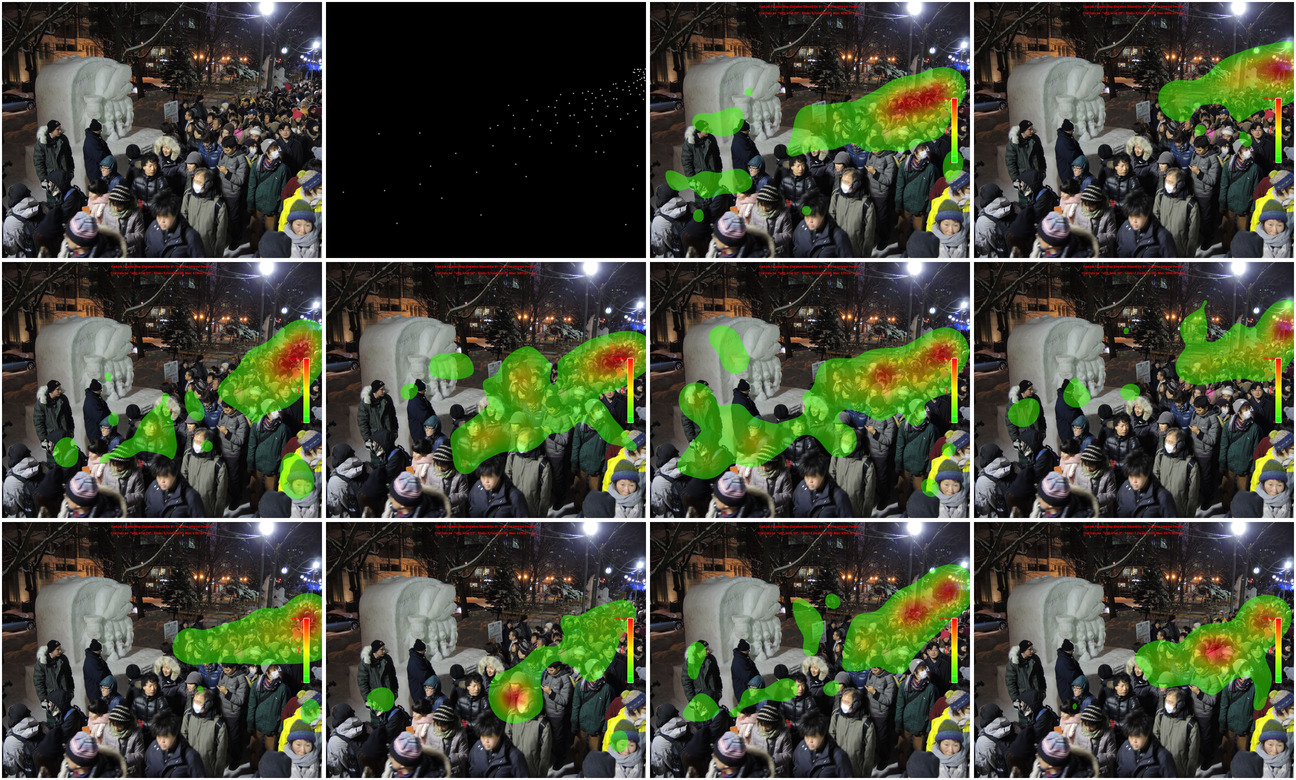}
    \caption{Image grid containing image 6, its annotation and fixation density maps  from all subjects. }
    \label{fig:image2}
\end{figure}

\begin{figure}
    \centering
    \includegraphics[width=0.9\linewidth]{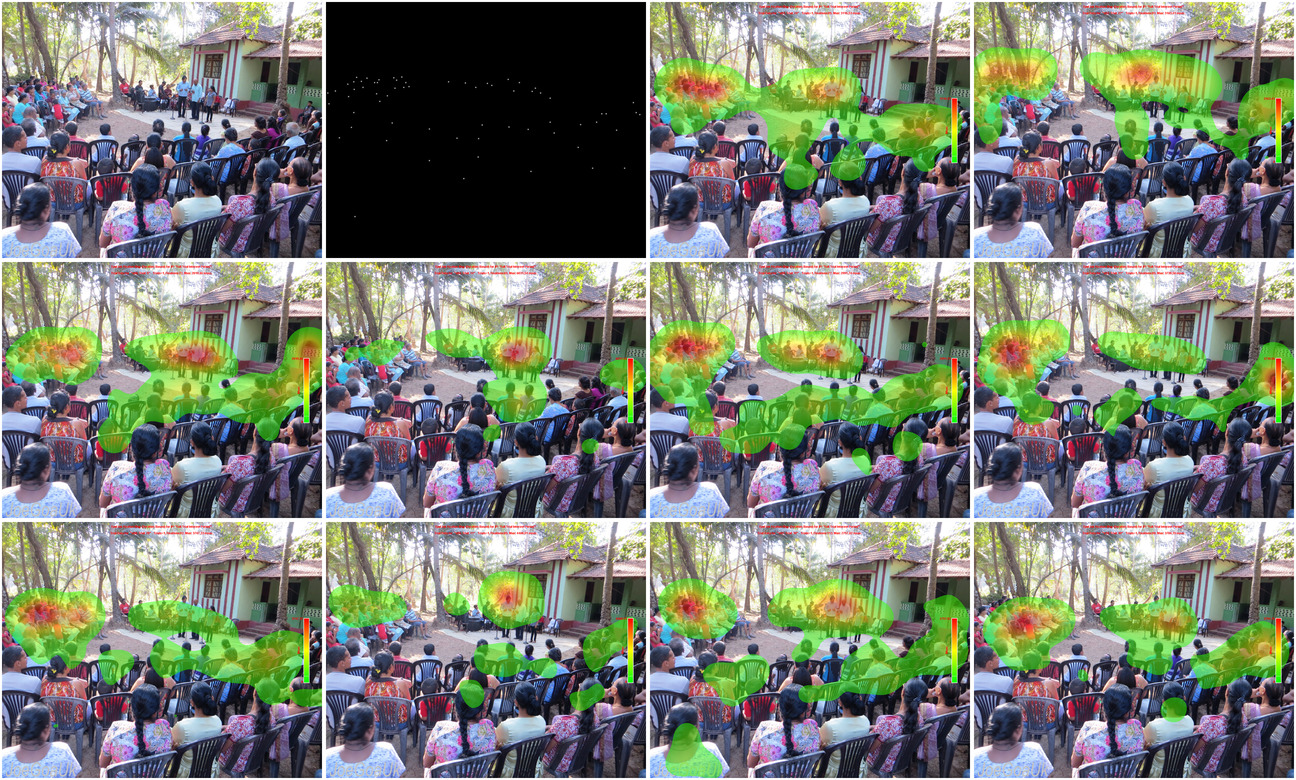}
    \caption{Image grid containing image 7, its annotation and fixation density maps  from all subjects. }
    \label{fig:image2}
\end{figure}

\begin{figure}
    \centering
    \includegraphics[width=0.9\linewidth]{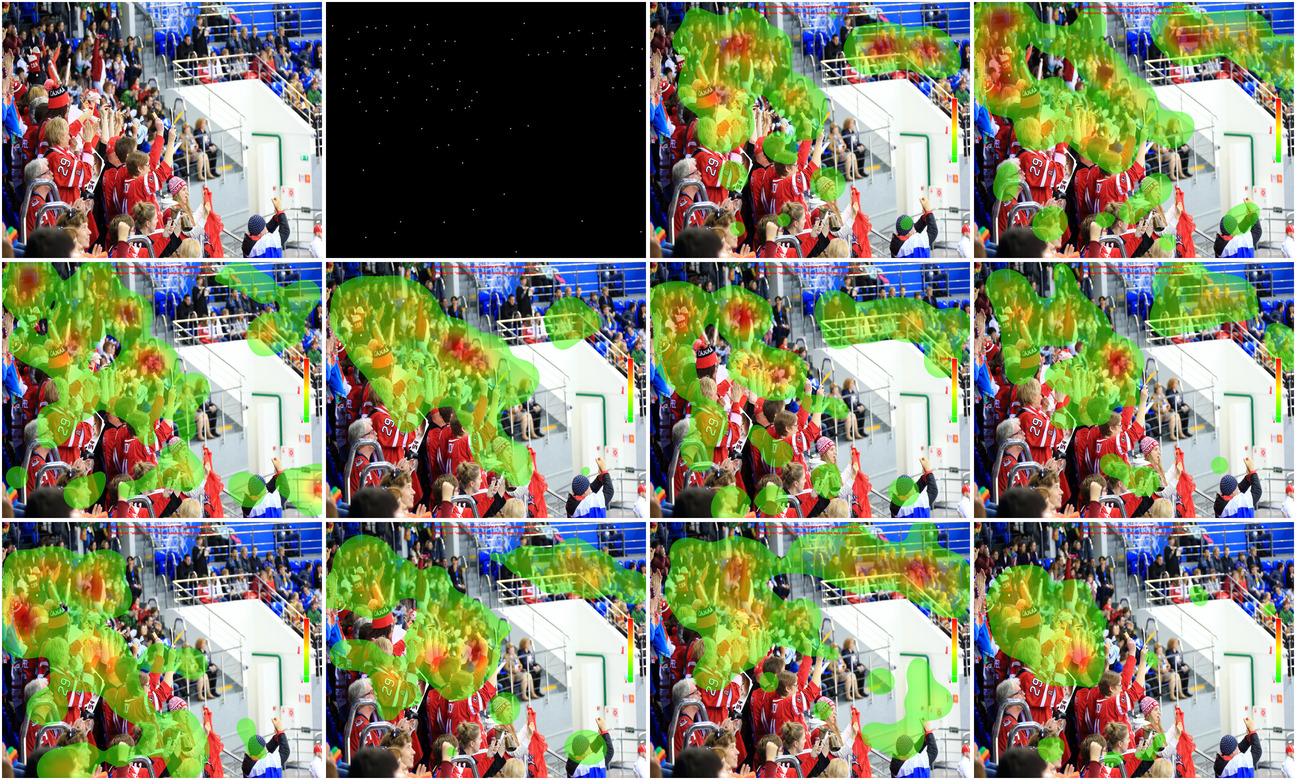}
    \caption{Image grid containing image 8, its annotation and fixation density maps  from all subjects. }
    \label{fig:image2}
\end{figure}

\begin{figure}
    \centering
    \includegraphics[width=0.9\linewidth]{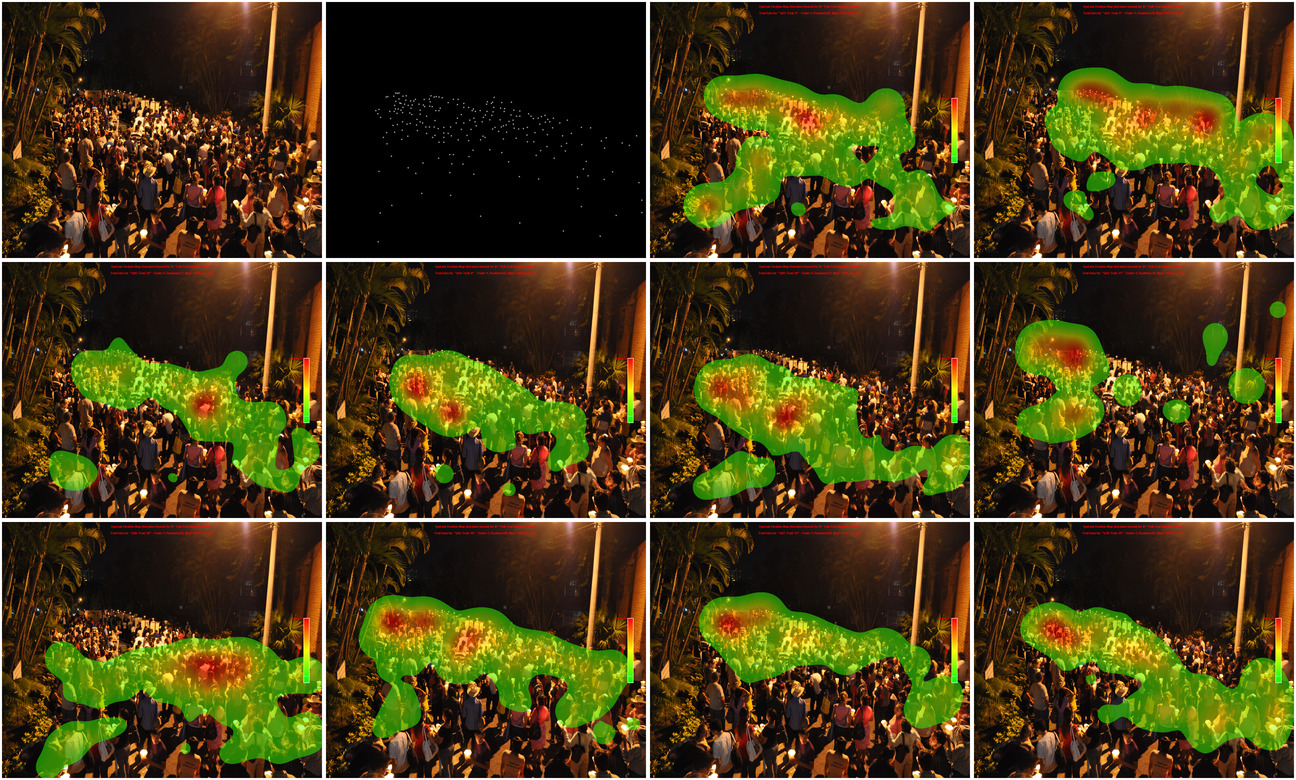}
    \caption{Image grid containing image 9, its annotation and fixation density maps  from all subjects. }
    \label{fig:image2}
\end{figure}

\begin{figure}
    \centering
    \includegraphics[width=0.9\linewidth]{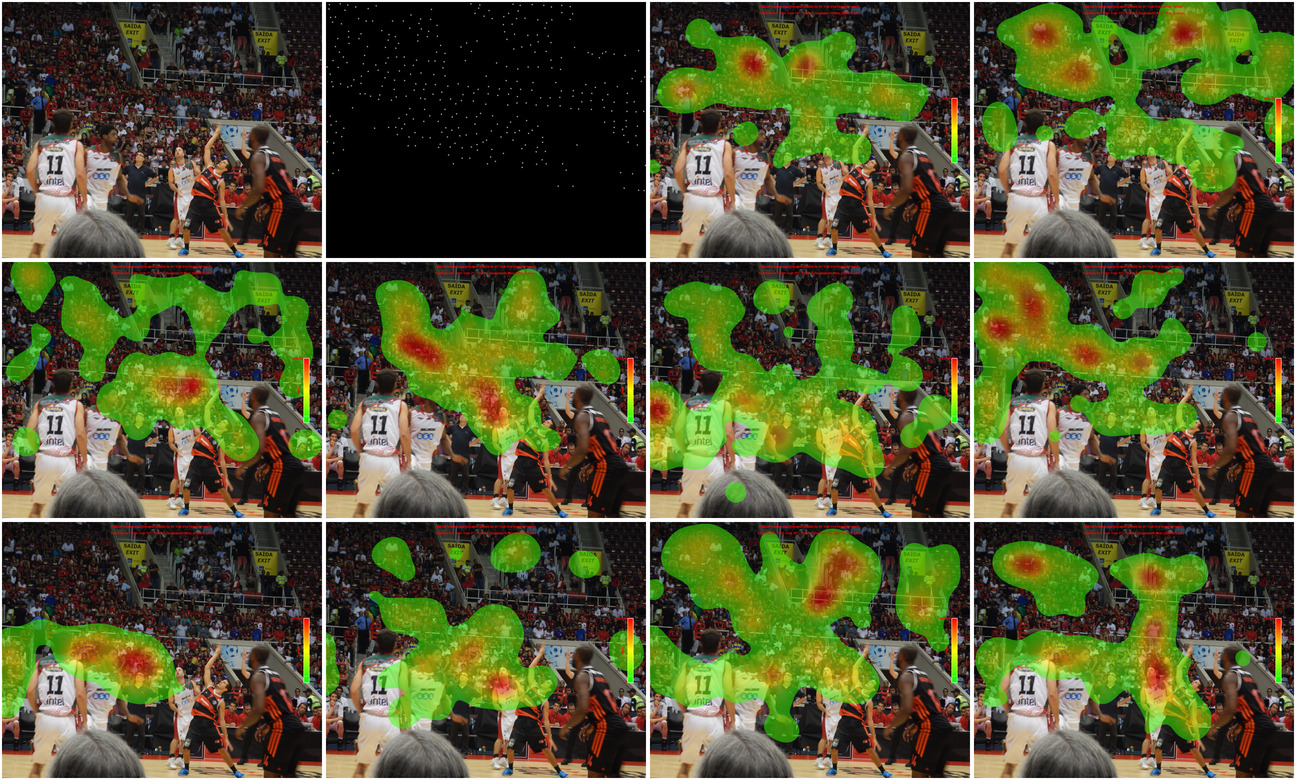}
    \caption{Image grid containing image 10, its annotation and fixation density maps  from all subjects. }
    \label{fig:image2}
\end{figure}

\begin{figure}
    \centering
    \includegraphics[width=0.9\linewidth]{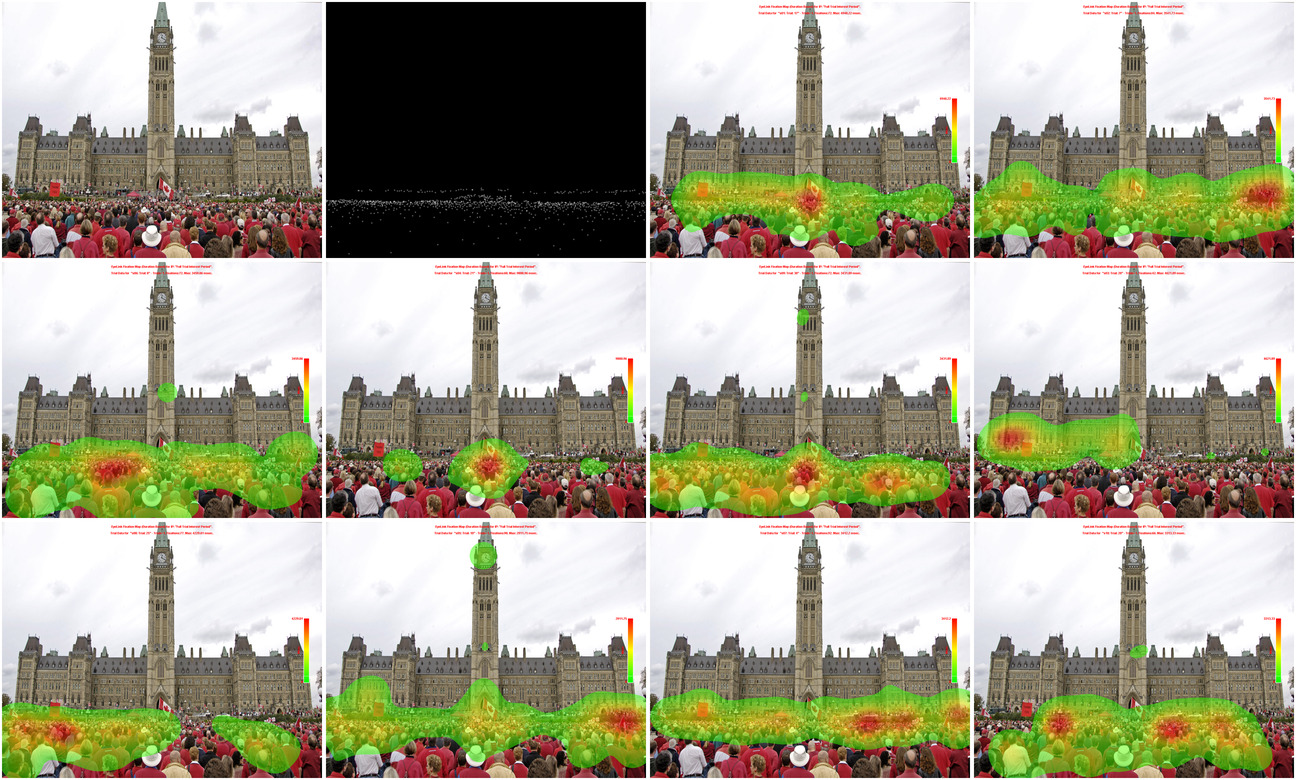}
    \caption{Image grid containing image 11, its annotation and fixation density maps  from all subjects. }
    \label{fig:image2}
\end{figure}

\begin{figure}
    \centering
    \includegraphics[width=0.9\linewidth]{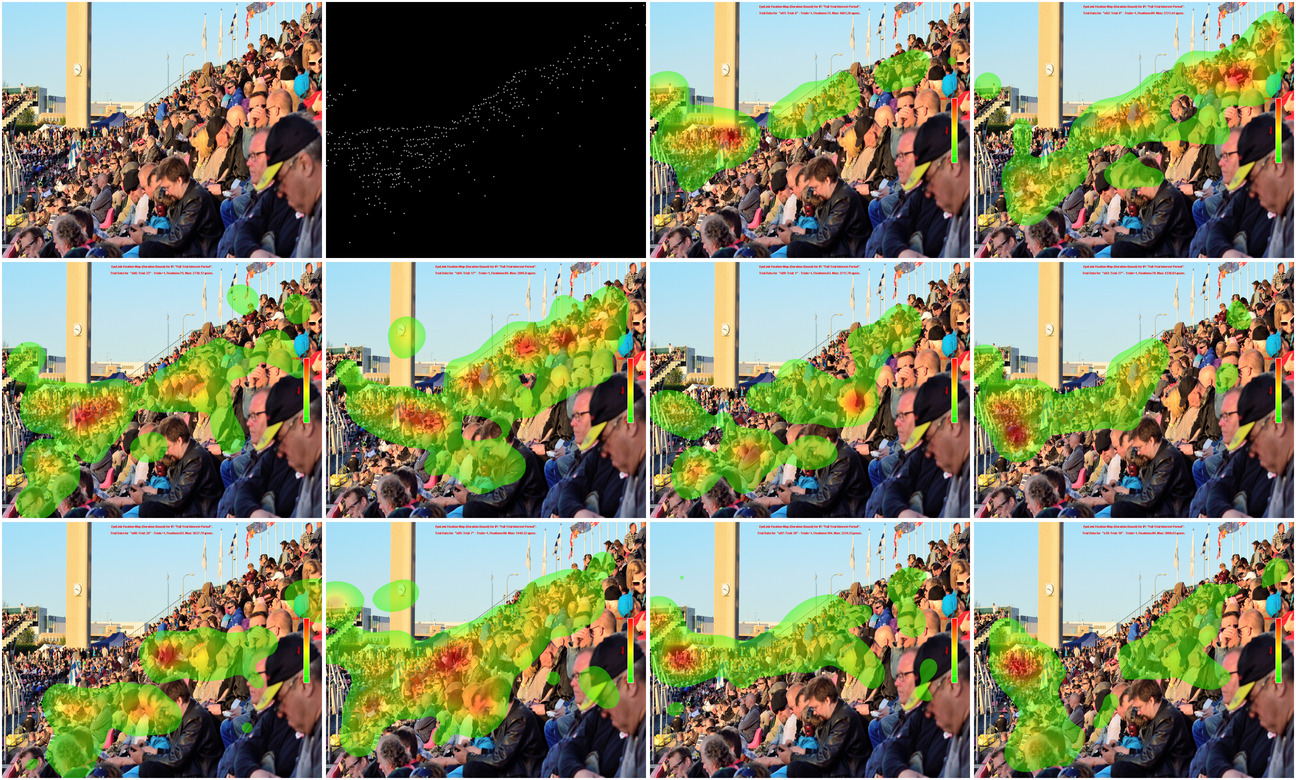}
    \caption{Image grid containing image 12, its annotation and fixation density maps  from all subjects. }
    \label{fig:image2}
\end{figure}

\begin{figure}
    \centering
    \includegraphics[width=0.9\linewidth]{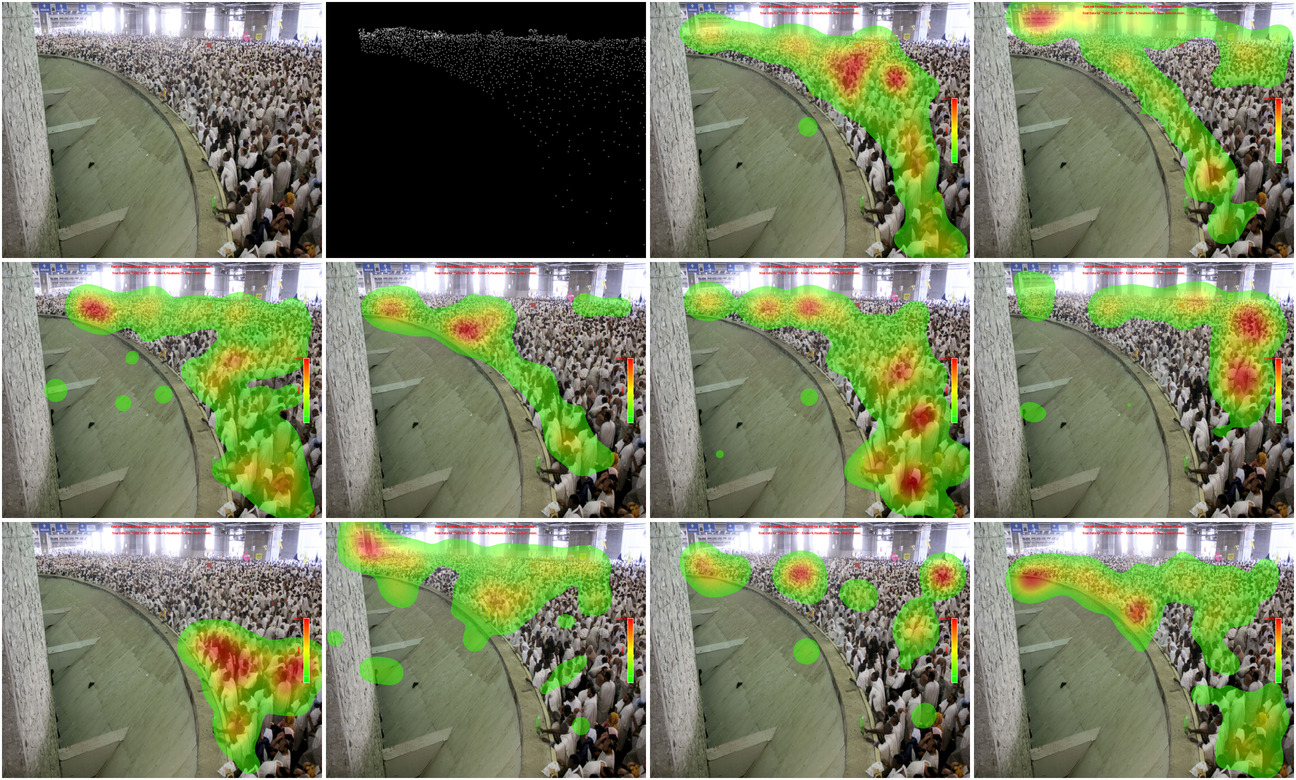}
    \caption{Image grid containing image 13, its annotation and fixation density maps  from all subjects. }
    \label{fig:image2}
\end{figure}
\begin{figure}
    \centering
    \includegraphics[width=0.9\linewidth]{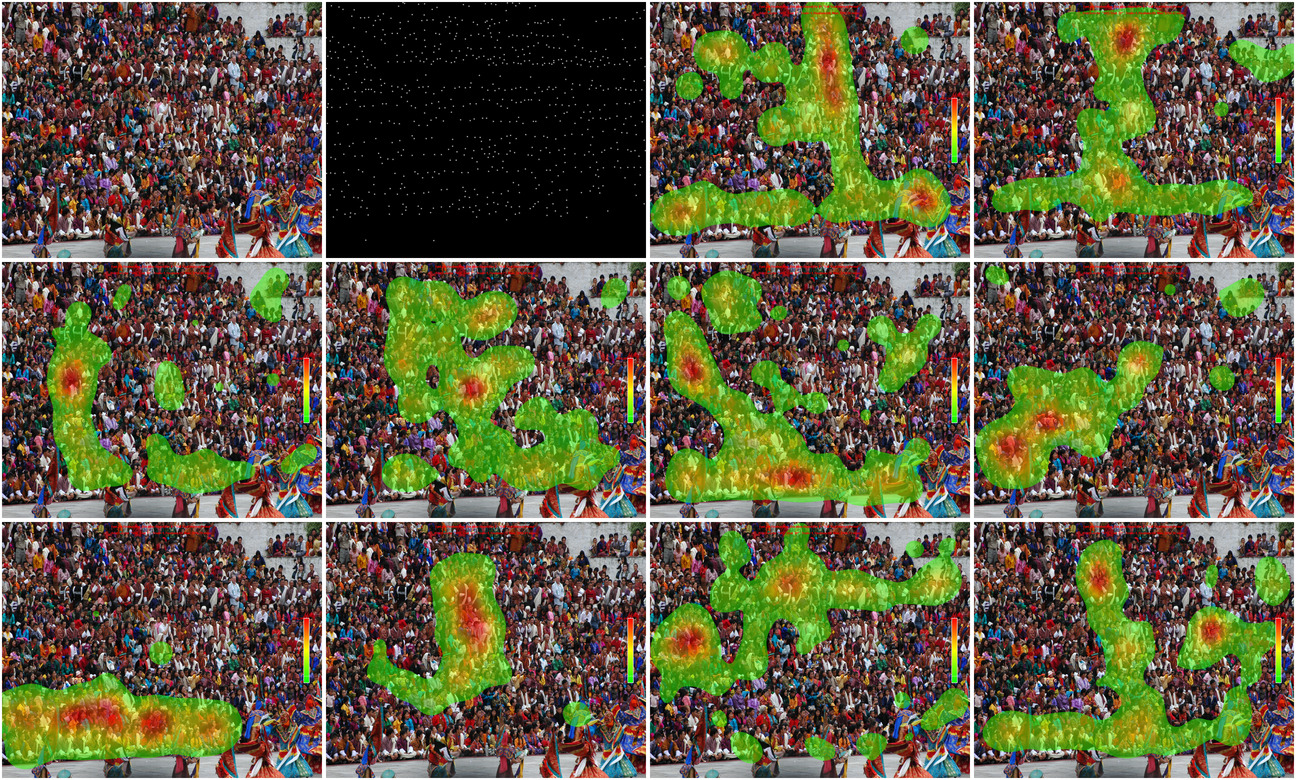}
    \caption{Image grid containing image 14, its annotation and fixation density maps  from all subjects. }
    \label{fig:image2}
\end{figure}

\begin{figure}
    \centering
    \includegraphics[width=0.9\linewidth]{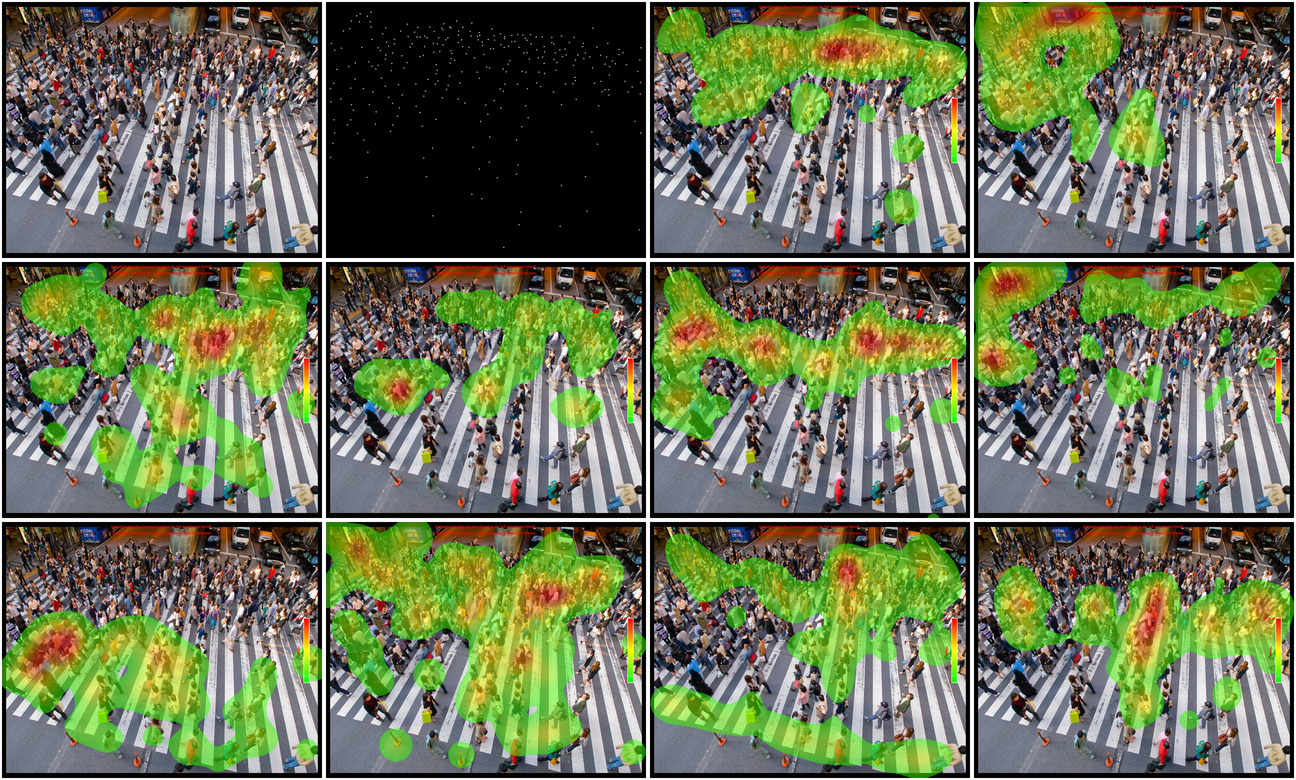}
    \caption{Image grid containing image 15, its annotation and fixation density maps  from all subjects. }
    \label{fig:image2}
\end{figure}

\begin{figure}
    \centering
    \includegraphics[width=0.9\linewidth]{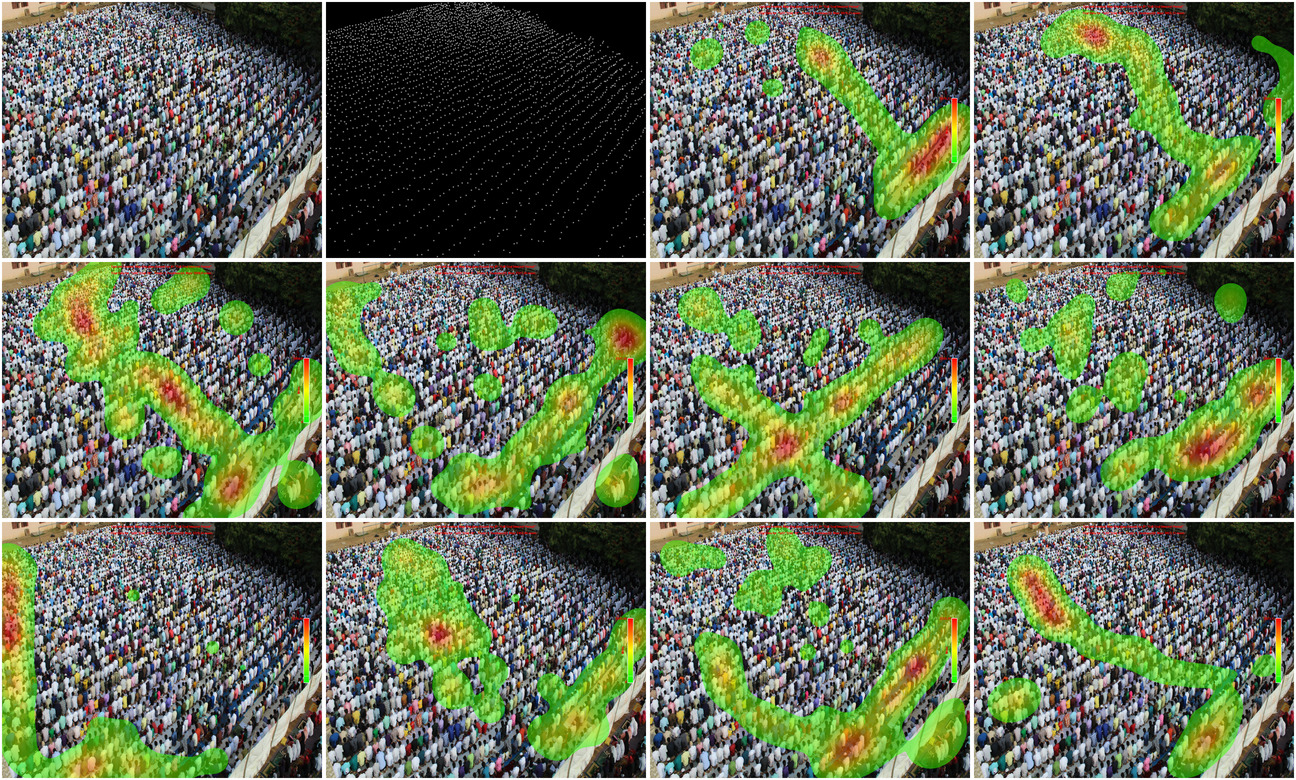}
    \caption{Image grid containing image 16, its annotation and fixation density maps  from all subjects. }
    \label{fig:image2}
\end{figure}

\begin{figure}
    \centering
    \includegraphics[width=0.9\linewidth]{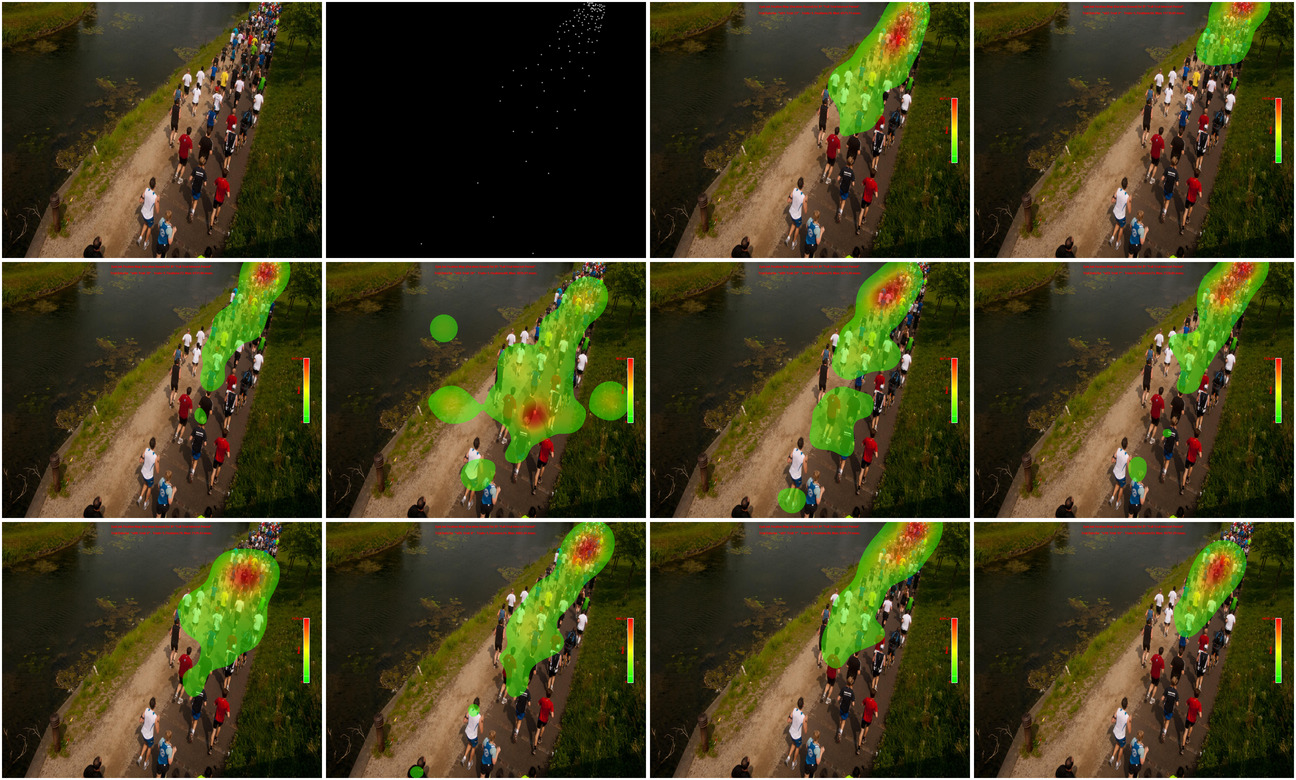}
    \caption{Image grid containing image 17, its annotation and fixation density maps  from all subjects. }
    \label{fig:image2}
\end{figure}

\begin{figure}
    \centering
    \includegraphics[width=0.9\linewidth]{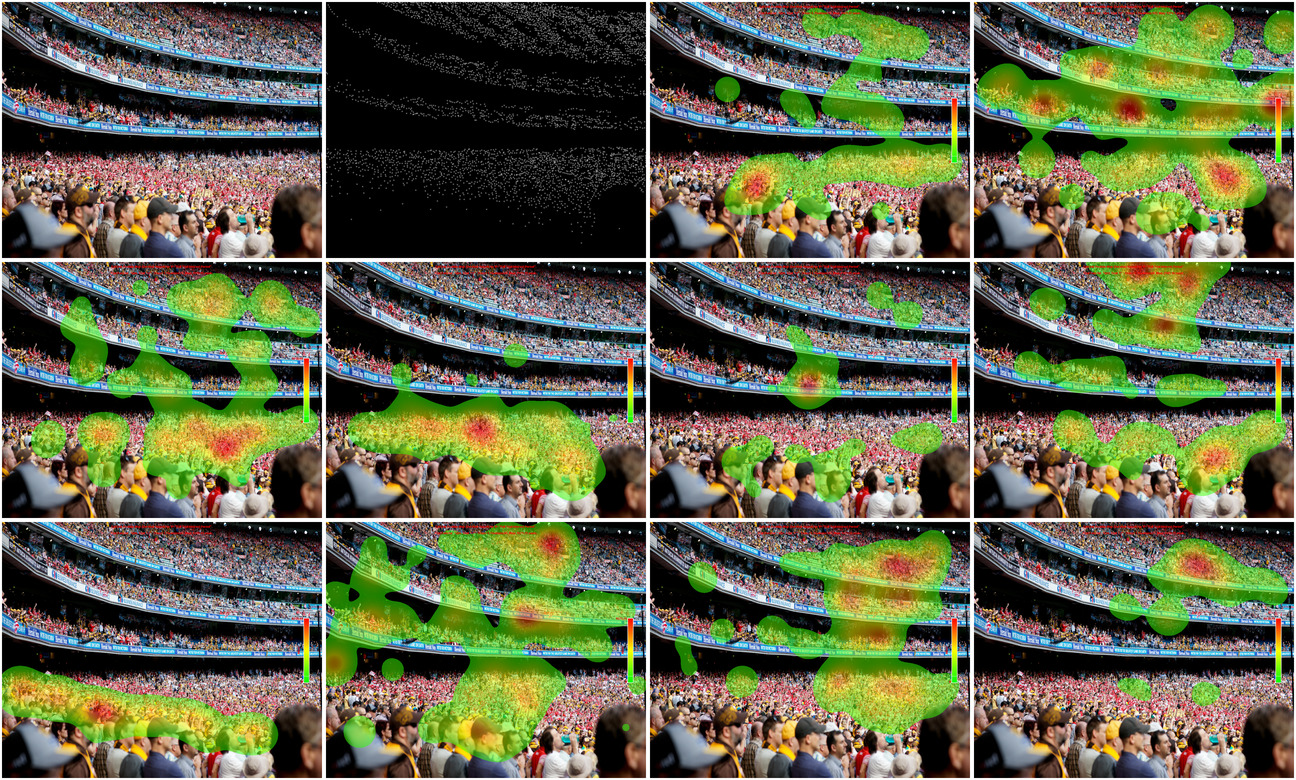}
    \caption{Image grid containing image 18, its annotation and fixation density maps  from all subjects. }
    \label{fig:image2}
\end{figure}

\begin{figure}
    \centering
    \includegraphics[width=0.9\linewidth]{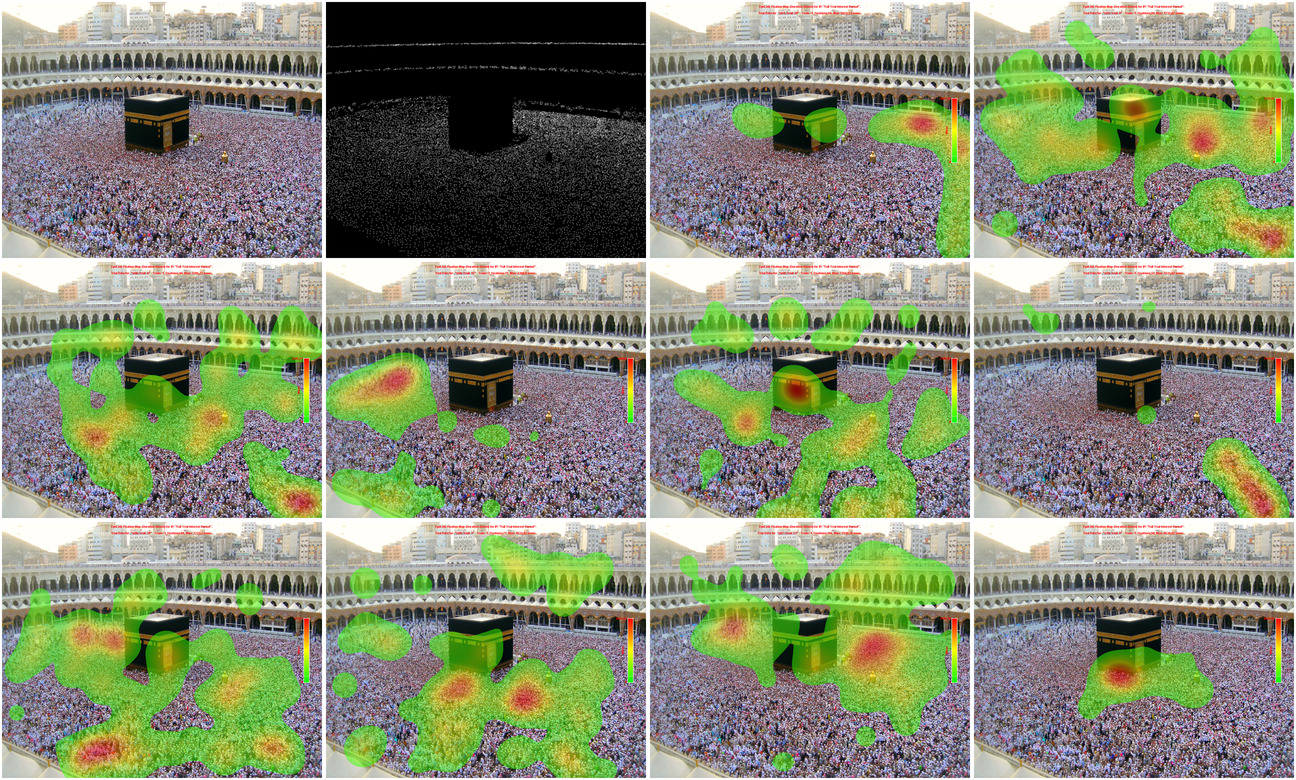}
    \caption{Image grid containing image 19, its annotation and fixation density maps  from all subjects. }
    \label{fig:image2}
\end{figure}

\begin{figure}
    \centering
    \includegraphics[width=0.9\linewidth]{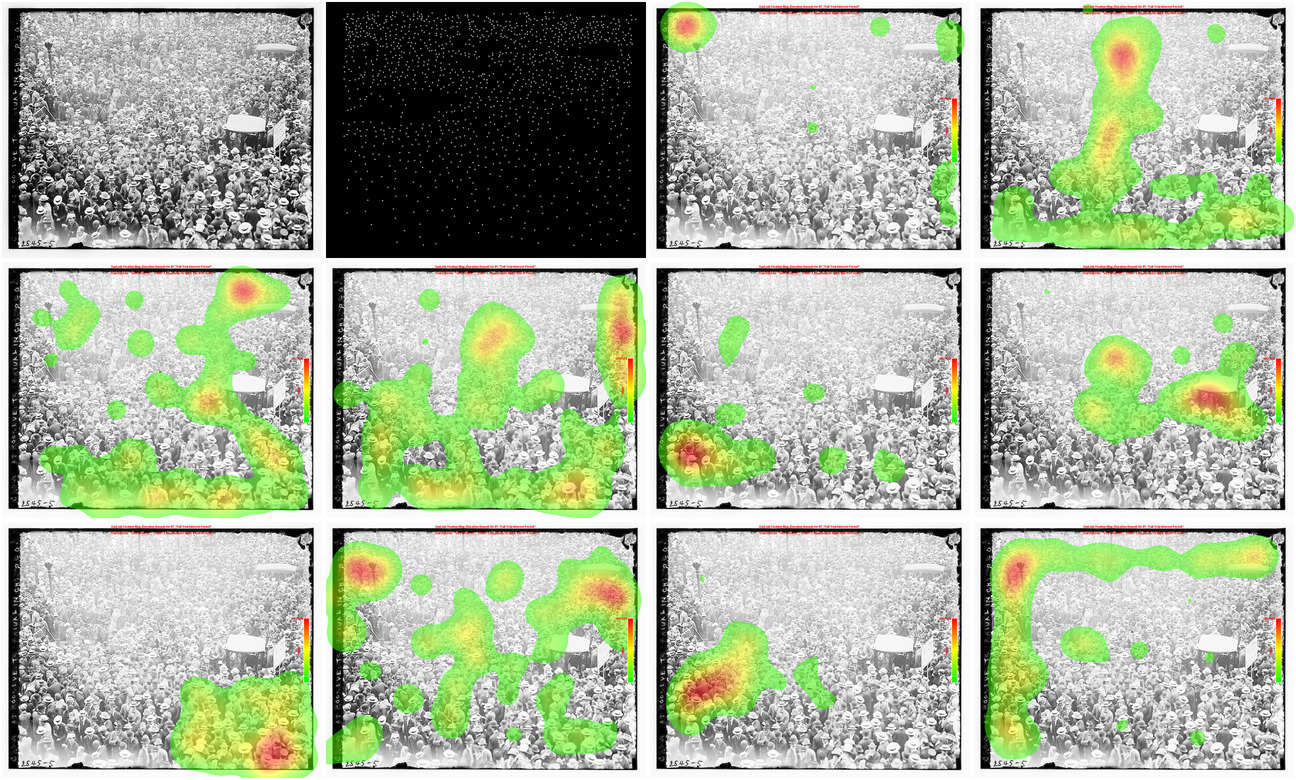}
    \caption{Image grid containing image 20, its annotation and fixation density maps  from all subjects. }
    \label{fig:image2}
\end{figure}

\begin{figure}
    \centering
    \includegraphics[width=0.9\linewidth]{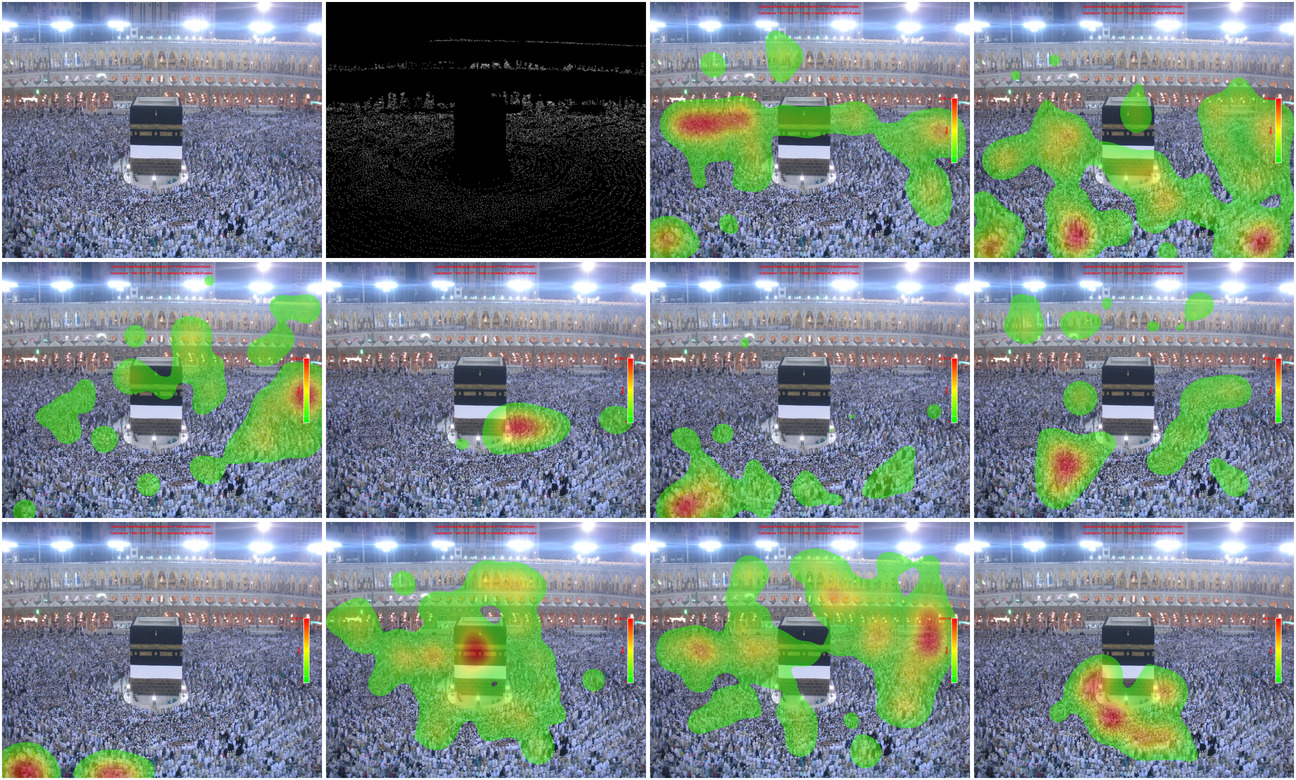}
    \caption{Image grid containing image 21, its annotation and fixation density maps  from all subjects. }
    \label{fig:image2}
\end{figure}

\begin{figure}
    \centering
    \includegraphics[width=0.9\linewidth]{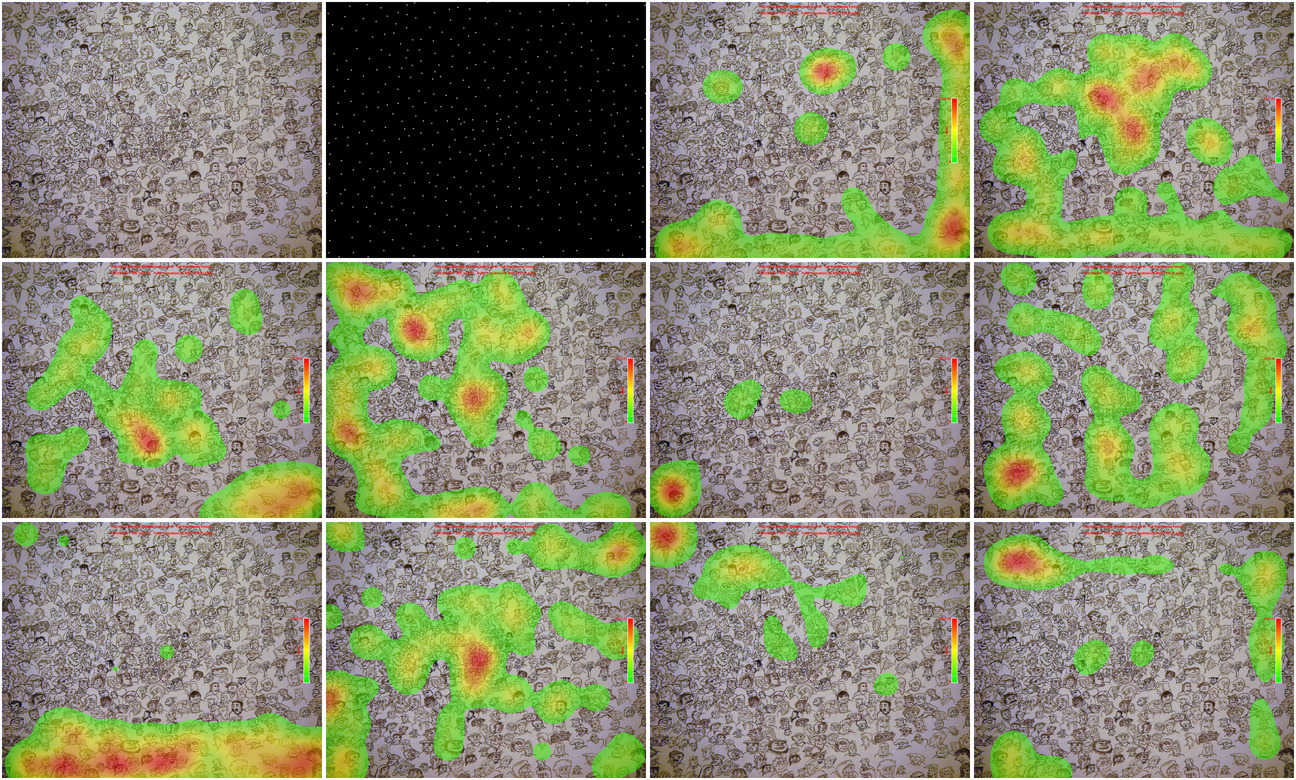}
    \caption{Image grid containing image 22, its annotation and fixation density maps  from all subjects. }
    \label{fig:image2}
\end{figure}

\begin{figure}
    \centering
    \includegraphics[width=0.9\linewidth]{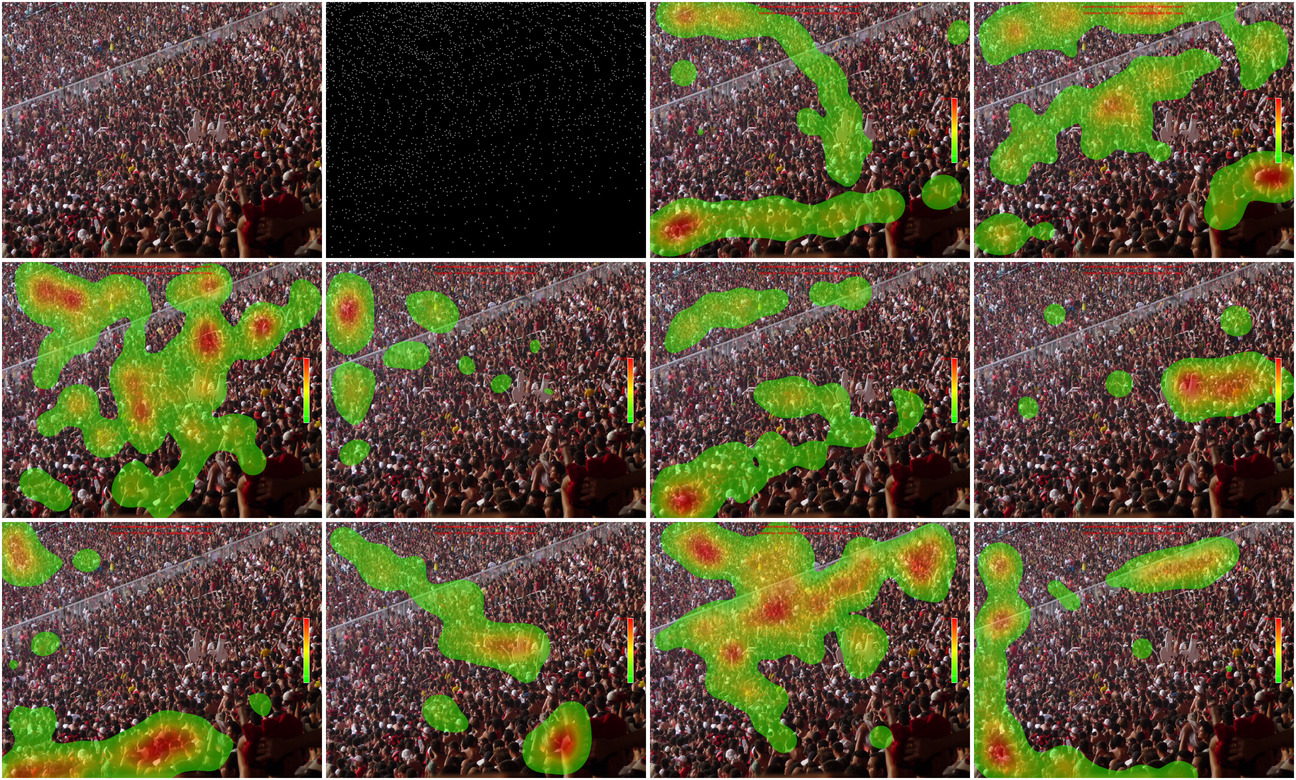}
    \caption{Image grid containing image 23, its annotation and fixation density maps  from all subjects. }
    \label{fig:image2}
\end{figure}

\begin{figure}
    \centering
    \includegraphics[width=0.9\linewidth]{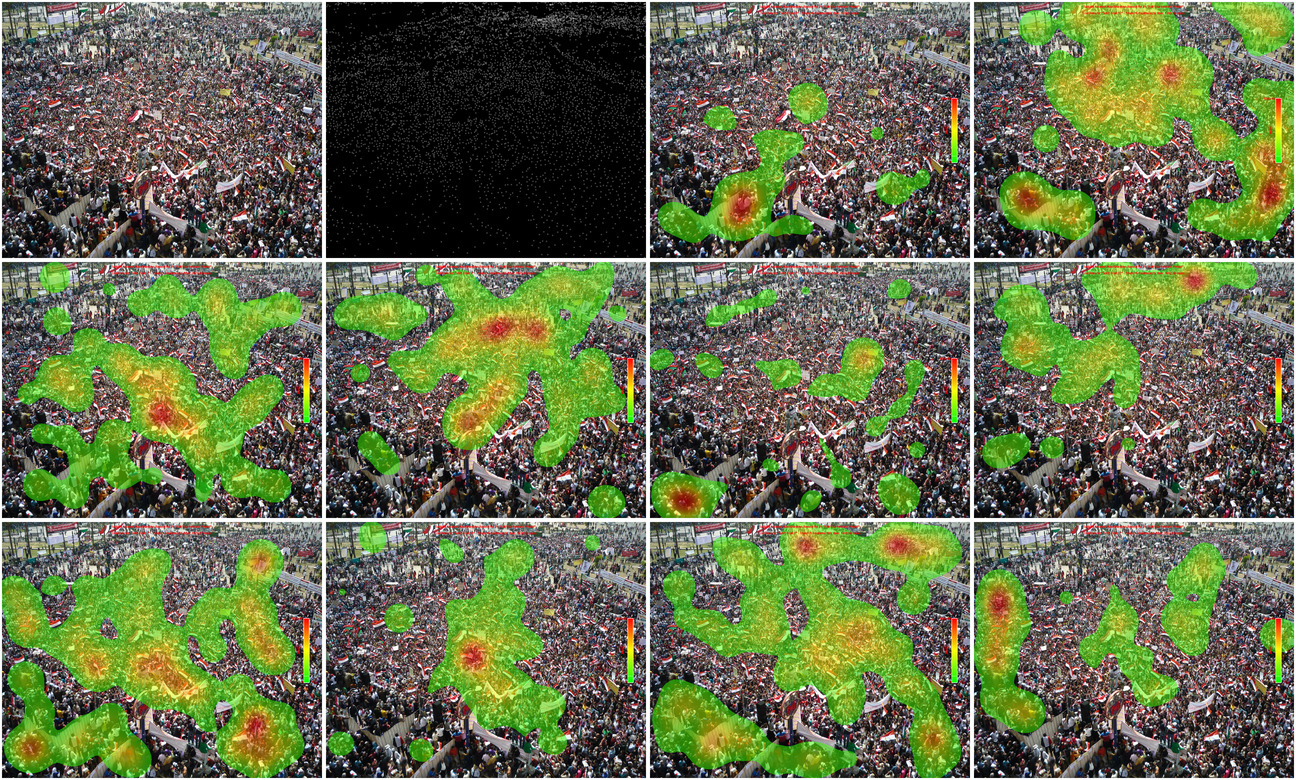}
    \caption{Image grid containing image 24, its annotation and fixation density maps  from all subjects. }
    \label{fig:image2}
\end{figure}

\begin{figure}
    \centering
    \includegraphics[width=0.9\linewidth]{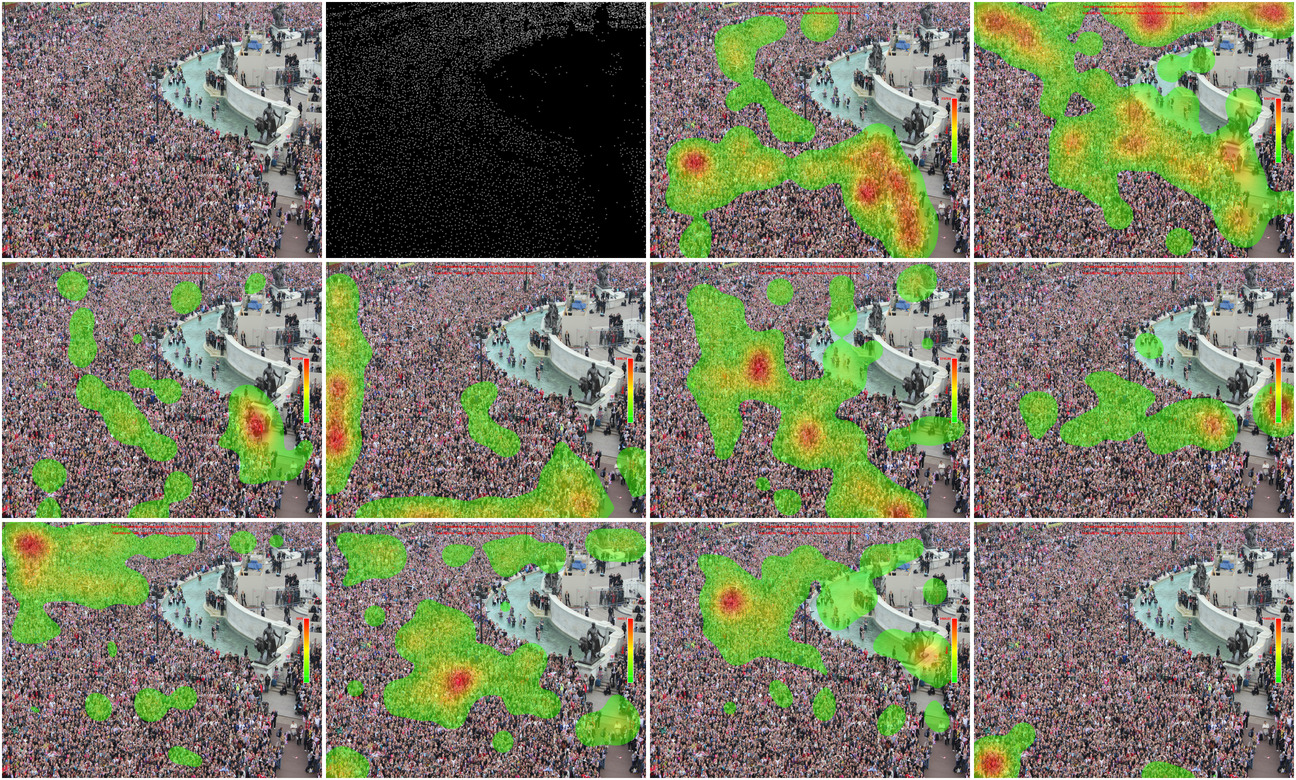}
    \caption{Image grid containing image 25, its annotation and fixation density maps  from all subjects. }
    \label{fig:image2}
\end{figure}

\begin{figure}
    \centering
    \includegraphics[width=0.9\linewidth]{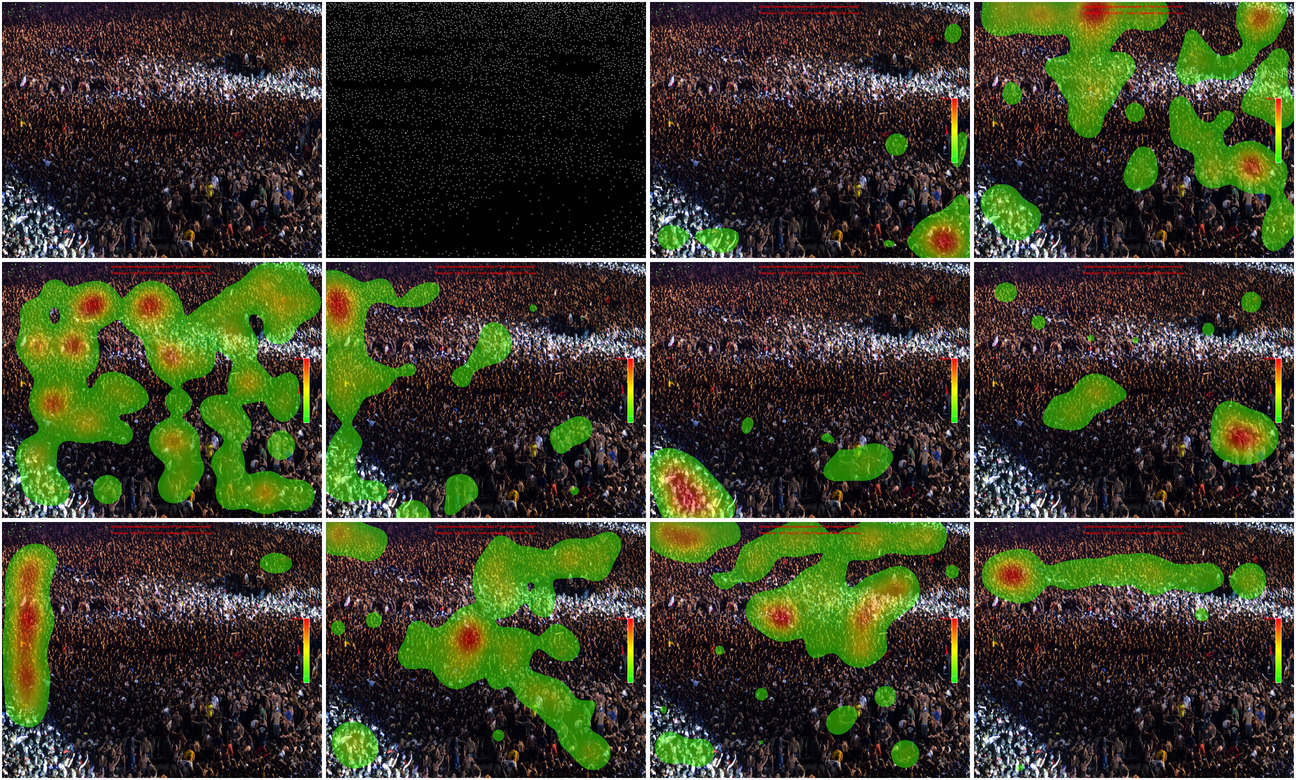}
    \caption{Image grid containing image 26, its annotation and fixation density maps  from all subjects. }
    \label{fig:image2}
\end{figure}

\begin{figure}
    \centering
    \includegraphics[width=0.9\linewidth]{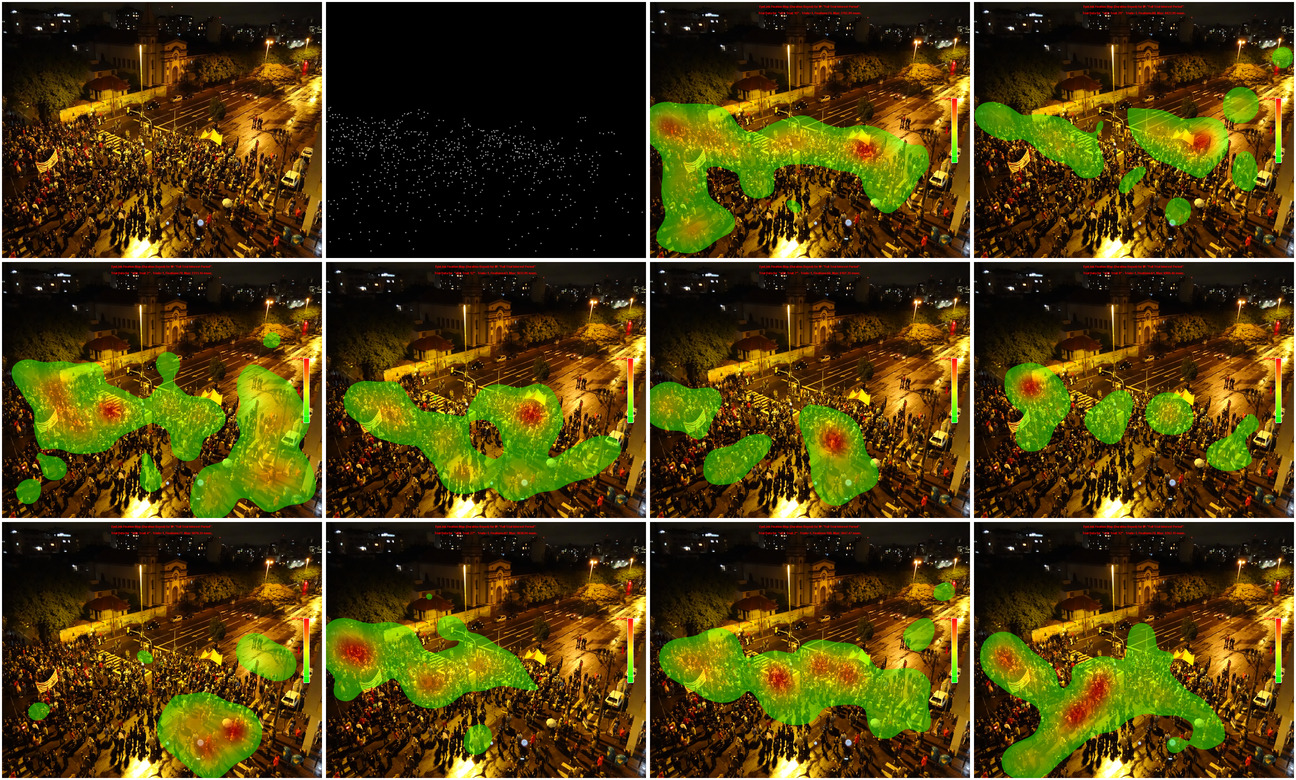}
    \caption{Image grid containing image 27, its annotation and fixation density maps  from all subjects. }
    \label{fig:image2}
\end{figure}

\begin{figure}
    \centering
    \includegraphics[width=0.9\linewidth]{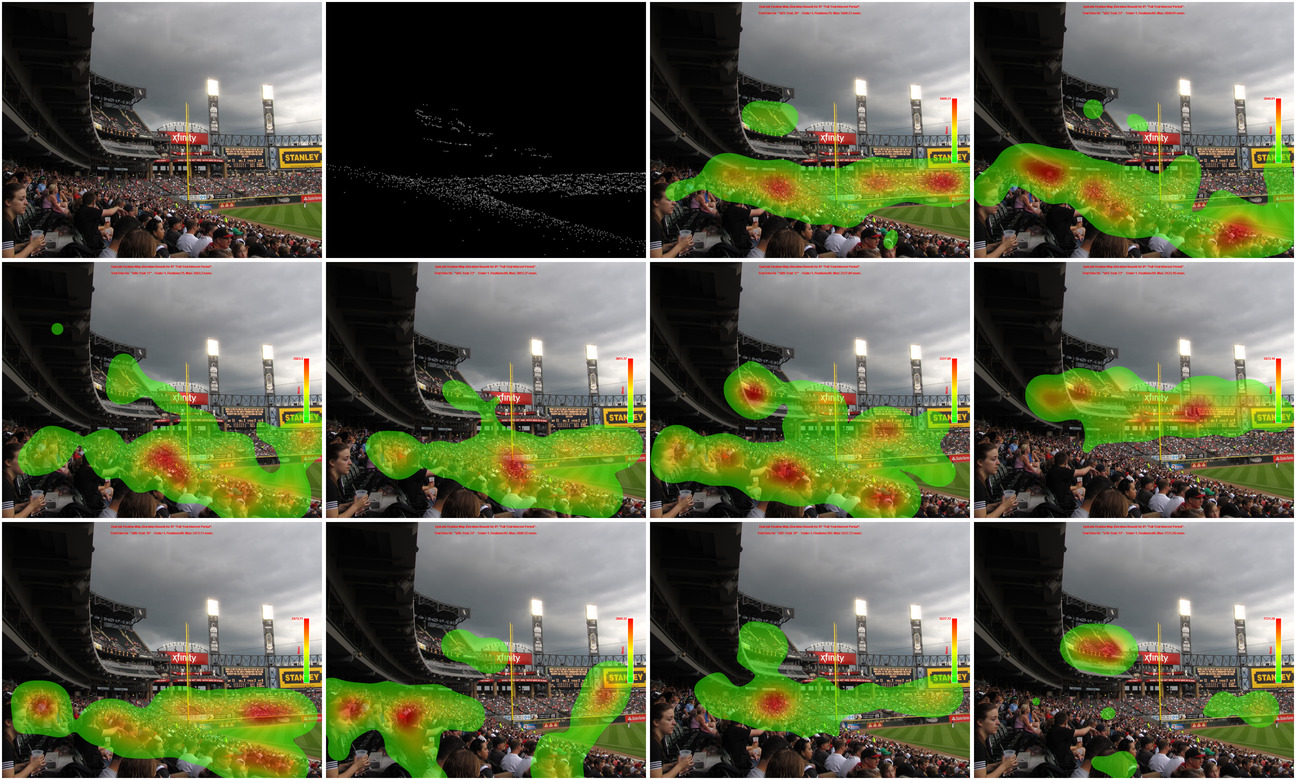}
    \caption{Image grid containing image 28, its annotation and fixation density maps  from all subjects. }
    \label{fig:image2}
\end{figure}

\begin{figure}
    \centering
    \includegraphics[width=0.9\linewidth]{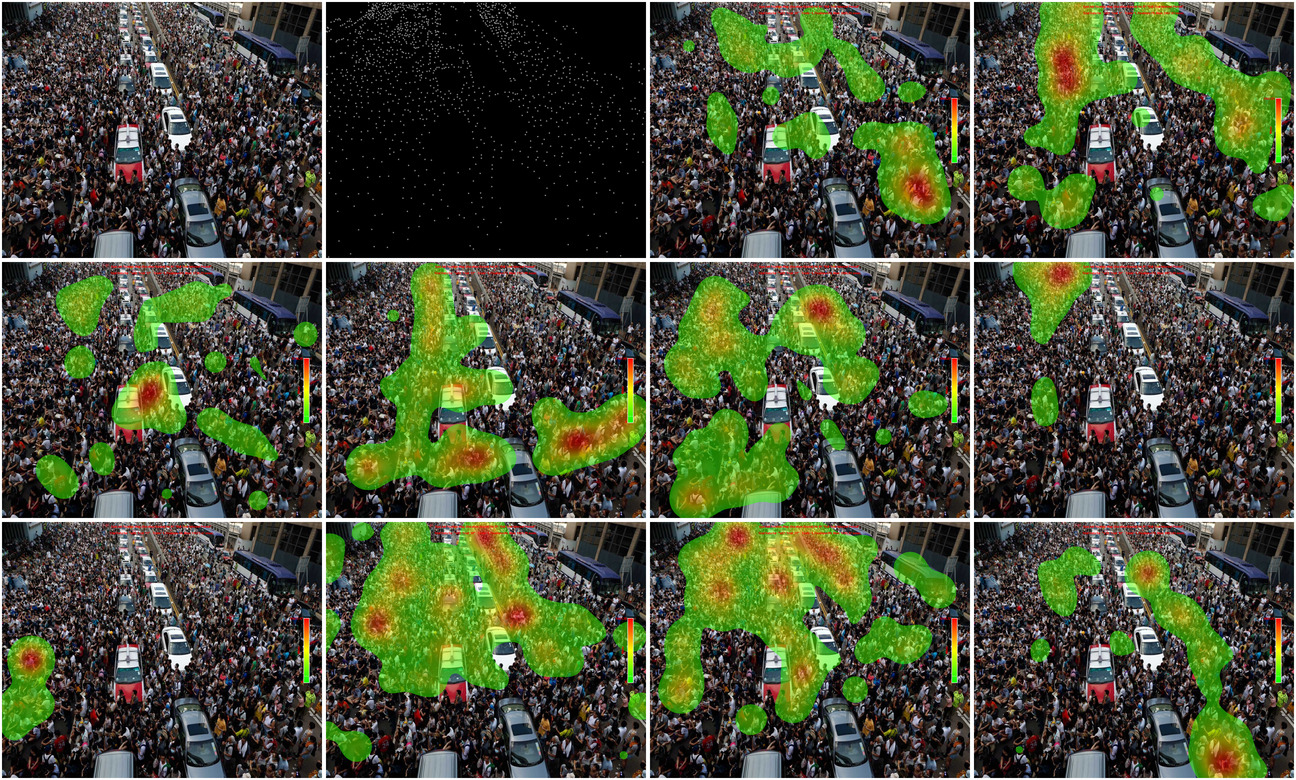}
    \caption{Image grid containing image 29, its annotation and fixation density maps  from all subjects. }
    \label{fig:image2}
\end{figure}

\begin{figure}
    \centering
    \includegraphics[width=0.9\linewidth]{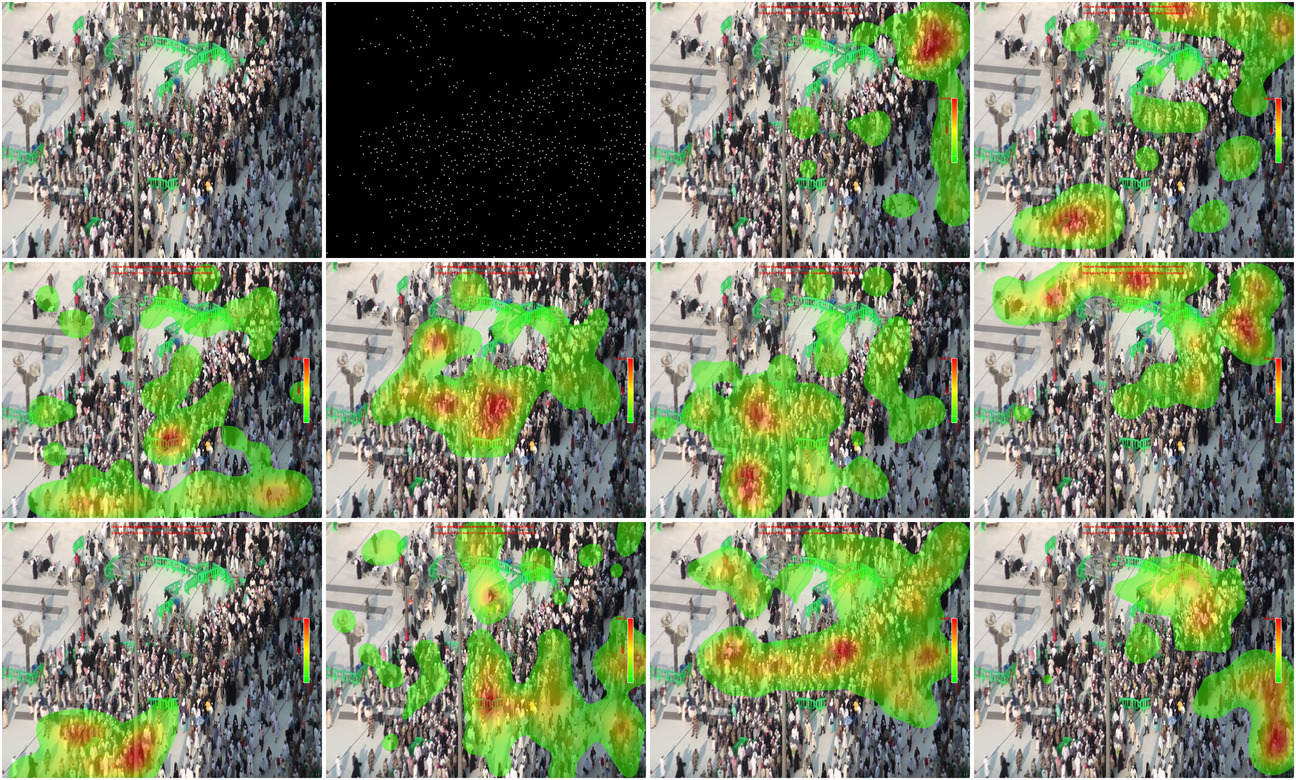}
    \caption{Image grid containing image 30, its annotation and fixation density maps  from all subjects. }
    \label{fig:image2}
\end{figure}
\fi 




    
  

\end{document}